\documentclass[conference]{IEEEtran}

\usepackage{array}
\usepackage{booktabs}
\usepackage{cite}
\usepackage{tikz}
\usepackage{amsmath}
\usepackage{amssymb}
\usepackage{textcomp}
\usepackage{multirow}
\usepackage{graphicx}
\usepackage[most]{tcolorbox}
\usepackage{colortbl}
\usepackage{algorithm,algpseudocode}
\usepackage{optidef}
\usepackage{makecell}
\usepackage{arydshln}
\usepackage{xspace}
\usepackage{tabularray}
\usepackage[caption=false,font=footnotesize]{subfig}
\usepackage{stfloats}
\usepackage{url} 
\usepackage{hyperref}

\newcommand{\sysname}{{PrivPair}\xspace}
\newcommand{\be}{\begin{equation}}
\newcommand{\ee}{\end{equation}}

\newtheorem{theorem}{Theorem}[section]

\newtheorem{assumption}{Assumption}

\begin{document}

\title{Privacy-Preserving Split Learning for Federated LLM Fine-Tuning}

\author{
	\IEEEauthorblockN{Heng Jin~~~~Chaoyu Zhang~~~~Hexuan Yu~~~~Wenjing Lou~~~~Y. Thomas Hou}
	\smallskip 
	\IEEEauthorblockA{\textit{Virginia Tech}, Arlington, VA, USA}
}

\maketitle

\begin{abstract}
Fine-tuning large language models (LLMs) on domain-specific data is essential for downstream adaptation. In many deployments, a participant cannot hold the complete model locally. This happens because the model owner keeps the full model proprietary, or because the participant lacks sufficient compute resources. Split Learning (SL) addresses this by partitioning the model between the participant and a server so that only a small portion runs locally. When the underlying data is additionally distributed across multiple institutions with privacy requirements, Federated Learning (FL) further enables collaborative training across participants by sharing only model updates instead of raw data. In this combined setting, each client transmits intermediate activations to the server, and for LLM fine-tuning, this exchange poses an inherent privacy paradox. The autoregressive nature of LLMs causes the transmitted activations to leak the input, and existing perturbation-based defenses are fundamentally ineffective in this setting. We address this leakage through a learned obfuscate-and-recover scheme that protects participants' private datasets while still allowing an independently deployable model to be trained on the server side. Experiments demonstrate that our approach achieves strong privacy protection with modest utility loss and system overhead, making split-based federated LLM fine-tuning practically viable.

\end{abstract}

\IEEEpeerreviewmaketitle

\section{Introduction}
\label{sec:introduction}

Large Language Models (LLMs) have demonstrated remarkable capabilities across a wide range of natural language processing tasks, covering both proprietary and open-weight models. Fine-tuning a pre-trained LLM on domain-specific data is essential for adapting it to downstream applications, with common paradigms including supervised instruction fine-tuning~\cite{wei2022flan} and reinforcement learning from human feedback~\cite{ouyang2022instructgpt}. However, the data needed for such fine-tuning is often private and distributed. Hospitals hold sensitive patient records, law firms hold confidential case files, and enterprises hold proprietary business data. Centralizing this data on a single server raises serious privacy concerns and may violate regulatory requirements, making collaborative fine-tuning without data sharing a practical necessity.

Two largely independent constraints make this collaborative setting harder in practice. First, a participant may be unable to hold the complete LLM locally. This happens when the model owner is unwilling to give participants access to the full set of proprietary model weights, or when the participant's hardware cannot load or train a full LLM, a common situation for institutions with limited accelerator infrastructure and for edge devices such as smartphones or embedded systems. Split Learning (SL)~\cite{SL_cite1} addresses this by partitioning the model between the participant and a server at a cut point, so that the participant executes only a lightweight portion locally while the server handles the bulk of the computation. Second, the fine-tuning data is often distributed across multiple institutions, each subject to its own privacy requirements. Federated Learning (FL)~\cite{FL_cite1, FL_survey_cite2, FL_survey_cite3, FL_survey_cite4} addresses this by allowing each institution to train on its local data and share only model updates, so raw data remains local. These two constraints frequently co-occur in practice, since the resource-constrained institutions and edge devices that need SL are often the same participants holding the private data that motivates FL. Federated split learning~\cite{FSL_survey__cite2, FSL_survey__cite3, FSL_survey_cite4} combines both techniques, letting each client execute only a lightweight portion of the model locally while still collaborating with other clients through federated aggregation.

However, applying SL to LLM fine-tuning poses a fundamental privacy challenge rooted in the autoregressive training objective. An LLM is trained to predict the next token given its input, so the input and label sequences are nearly identical, and leaking the labels is effectively equivalent to leaking the input. Assigning both the input-side and the output-side of the model to the client keeps raw inputs and labels local, yet SL still requires the client to transmit intermediate activations (smashed data) to the server. Because of the autoregressive nature of LLMs, a white-box server can reconstruct labels from these activations and thereby recover the original input. This leakage path makes state-of-the-art defenses that only perturb smashed data insufficient for LLM fine-tuning. 

\begin{figure}[t]
    \centering
    \includegraphics[width=\linewidth]{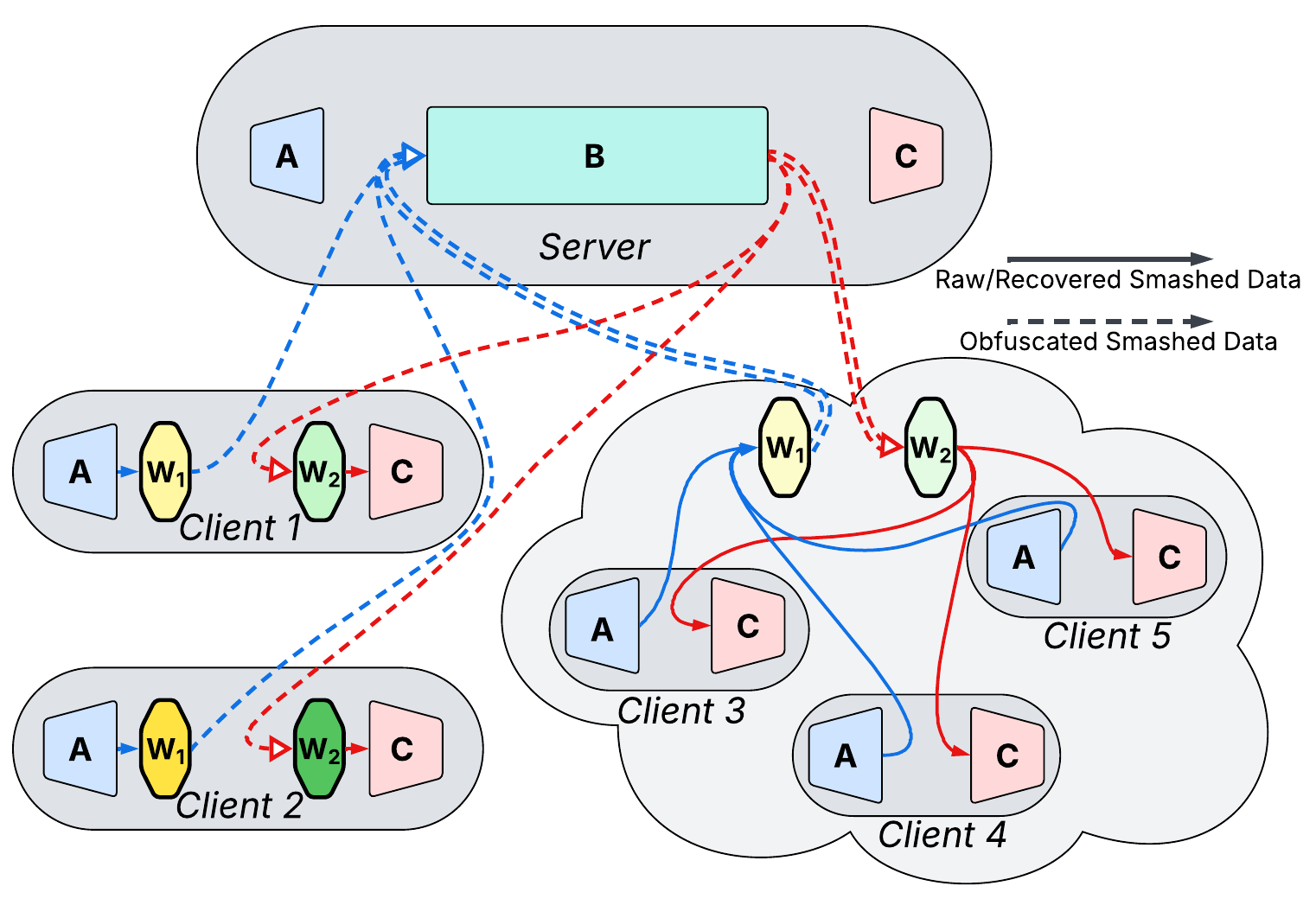}
    \caption{Overview of the FL scenario with multiple clients and a server exchanging smashed data. Line colors are used only to improve visual distinguishability. Clients 3, 4, and 5 are mutually trusted and share a single adapter pair. The raw smashed data produced by $A$ is obfuscated by $W_1$ before being sent to the server, and the server's response is recovered by $W_2$ before being consumed by $C$. After aggregation and global model synchronization, every client and the server hold the same $A$ and $C$, while each client (or trust group) retains its own distinct adapter pair. }
    \label{fig:fl_plane}
\end{figure}

To address this challenge, we propose \sysname, a framework that makes split-based federated LLM fine-tuning practical under our threat model. The key insight behind \sysname is to deploy lightweight adapters on the client to obfuscate the smashed data sent to the server, so that these activations are no longer compatible with the server's decoder. After the smashed data is processed by the server and returned to the client, a recovery adapter maps it back to a representation compatible with the client-side model, preserving training utility. Figure~\ref{fig:fl_plane} illustrates the resulting scenario.

\noindent\textbf{Contributions.} Our contributions are summarized as follows:

\begin{itemize}
    \item We propose \sysname, a client-side obfuscate-and-recover mechanism with two lightweight adapters that mitigates smashed data leakage while preserving trainability.
    \item \sysname allows the server to retain a complete, directly deployable model after training, preserving the standard FL deployment workflow.
    \item We evaluate \sysname against reconstruction attacks and baseline defenses, showing strong privacy protection, modest utility loss, and modest client-side computation and memory overhead on devices with limited GPU memory.
\end{itemize}

\noindent Our artifact is available at \url{https://github.com/hengvt/PrivPair_arXiv}.

\section{Related Work}\label{background}

\subsection{Federated Learning and Split Learning}
FL~\cite{FL_cite1} and SL~\cite{SL_cite1} represent two independent lines of work for collaborative model training. FL addresses data privacy by enabling multiple clients to collaboratively train a shared model while keeping raw data local, exchanging only model updates. SL addresses computational efficiency by partitioning a neural network at a cut point and offloading the major portion of the model to the server, so that resource-constrained clients need only execute a lightweight head locally. Federated split learning~\cite{FSL_cite1} combines both ideas. It adopts the model-partitioning approach of SL to reduce client burden together with the multi-client aggregation protocol of FL to enable collaborative training without sharing raw data. In LLM fine-tuning, the client may also retain the output-side component of the model to keep labels and loss computation local, which raises the risk of label leakage from the smashed data exchanged at the cut points.

\subsection{Data Reconstruction Attacks}
Despite the privacy-preserving objective of FL and SL, prior work has identified multiple attack surfaces, including property inference attacks~\cite{property_cite1}, membership inference attacks~\cite{membership_cite1, membership_cite2}, and data reconstruction attacks (DRA)~\cite{DRA_cite1, DRA_cite2, DRA_cite3, DRA_cite4}. Among these, DRA is particularly concerning, as it directly targets the information exposed during collaborative training. If the server can reconstruct the client's raw training data from the information it receives, the central privacy motivation of FL is undermined. Existing DRA methods in SL target both the forward and backward communication paths. EIA~\cite{EIA_attack} targets the smashed data transmitted from the client to the server. Given the intermediate hidden states produced by the client-side model, EIA trains an inverse mapper or performs optimization to find input tokens whose representations match the observed smashed data. DLG~\cite{DLG_attack} initializes dummy inputs and labels, then minimizes the distance between their gradients and the observed gradients to recover the original training sample. TAG~\cite{TAG_attack} extends DLG to Transformer-based language models with an additional regularization term, yielding higher text recovery rates. LAMP~\cite{LAMP_attack} further improves upon TAG by introducing a language model prior to guide the search toward natural text, combined with discrete token reordering to recover more accurate token sequences. BiSR~\cite{CCS_unveiling} proposes a bidirectional attack that jointly exploits both the forward and backward paths, combining smashed data inversion with gradient matching to achieve stronger reconstruction in the LLM SL setting. This line of work motivates our focus on the smashed data and gradients exchanged at the split points.

BiSR also shows that label leakage remains severe even when both the input-side and the output-side of the model reside on the client. They attribute this persistence to two fundamental obstacles. The first is the autoregressive nature of LLMs. SL requires the client to transmit intermediate activations to the server, which can then exploit its white-box copy of the model to reconstruct the labels from these activations and thereby infer the original input. The second is the not-too-far property of LLM fine-tuning. Even if the client withholds its fine-tuned parameters from the server, the parameter differences introduced by fine-tuning are sufficiently small that the server can still use its copy of the pre-trained model to accurately predict the labels. Under the conventional FL objective of producing a complete, deployable model on the server, this second obstacle becomes even more severe, because the server can hold the same model as the clients and label inference becomes substantially easier.

\subsection{Privacy Protection Methods}
Existing defenses for SL include perturbation-based methods and training-time regularizers, both aiming to prevent the server from recovering the client's private input. Perturbation-based methods differ in where the noise is applied. Embedding $\varepsilon$-Privacy~\cite{DP_attack} applies noise at the embedding layer output, perturbing the token embeddings to satisfy a relaxed local differential privacy condition~\cite{DP_attack_2}. Smashed-Data DP~\cite{DP_forward_attack, DP_forward_attack_2, DP_forward_attack_3} applies Laplace or Gaussian noise directly to the smashed data transmitted to the server. NoPeek~\cite{NoPeak} takes a training-time approach, adding a distance correlation regularization term to the training loss to minimize the statistical dependence between the smashed data and the original input. These methods share a common limitation in the LLM fine-tuning setting. The same representation sent to the server must both hide the private input and remain directly useful for the downstream model. This coupling forces privacy protection and trainability to compete through a single smashed-data representation, motivating our obfuscate-and-recover design.
\section{Threat Model}
\label{sec:threat_model}

\subsection{System Model}

\begin{figure}[t]
    \centering
    \includegraphics[width=\linewidth]{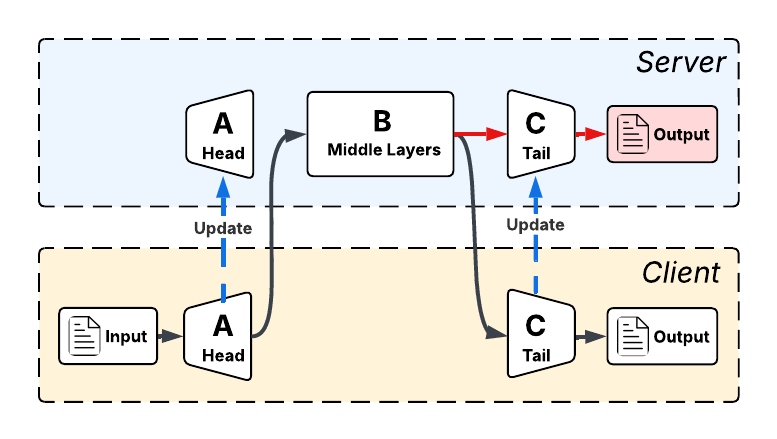}
    \caption{The forward path of the baseline split FL system for LLM fine-tuning (one client shown as an example). The LLM is split into three components $A$, $B$, and $C$, with $A$ and $C$ residing on the client and $B$ on the server. Note that since the server holds a copy of $C$, it can apply $f_C$ directly to $s_B$ to predict the labels, exactly as the client would.}
    \label{fig:base_model}
\end{figure}

We consider a split FL system in which an LLM is partitioned into three components and distributed across clients and a server, as illustrated in Figure~\ref{fig:base_model}. Each client holds the head component $A$ and the tail component $C$, while the server holds the middle component $B$. During fine-tuning, clients transmit the intermediate activations (smashed data) $s_A$ to the server at the first cut point, receive the processed activations $s_B$ back at the second cut point, and exchange the corresponding gradients in the backward pass. The server aggregates client updates via FedAvg at the end of each round.

\noindent\textbf{Adversary's Capabilities.}
We assume the server acts as an \textit{honest-but-curious} adversary. It follows the prescribed protocol and does not deviate from the training procedure, but it attempts to infer clients' private training data from all information it legitimately receives. The adversary has the following capabilities: (i)~\textit{Access to smashed data and gradients.} The server has full access to the smashed data and cut-point gradients exchanged by all clients at every training step. (ii)~\textit{White-box access to the model.} The server has complete white-box knowledge of the model architecture and current parameters, including both the server-side and the client-side portions. This knowledge excludes the client-side adapters introduced by our defense, which remain local to each client.

In conventional FL, clients typically perform multiple local training steps before uploading their updates, so the server has no knowledge of the intermediate client-side weights during those steps. To construct a stronger threat model, we grant the server knowledge of the client-side weights after every local training step. This ensures the server maintains full and current knowledge of the client-side model weights throughout the entire fine-tuning process without changing the training protocol itself.

\noindent\textbf{Defender's Capabilities.}
The defender has the following capabilities: (i)~\textit{Arbitrary auxiliary dataset.} The defender may use an arbitrary text dataset that contains no private information. This dataset is independent of the private fine-tuning data and does not need to match its task, domain, or distribution. It may be obtained from any public or synthetic source. (ii)~\textit{Client-side adapters.} The defender may train two lightweight adapters on the client side. These adapters remain local at all times, are never disclosed to the server at any stage of training or deployment, and can be freely discarded once training is complete.

\noindent\textbf{Objectives.}
The adversary aims to reconstruct as much of the original text from the client's private fine-tuning dataset as possible. The defender aims to prevent such reconstruction while ensuring that the fine-tuning process remains effective, producing a model of comparable quality to one trained without any privacy protection. The defender also requires the server to retain a complete, directly deployable model after training, consistent with the standard FL objective.

\subsection{Scope}
A separate class of attacks targets the model updates uploaded by clients at the end of each round. These gradient inversion attacks are independent of both SL and the LLM setting, and have been extensively studied in the standard FL literature, with well-established defenses such as secure aggregation~\cite{bonawitz2017secureagg,bell2020secureagg} and differential privacy~\cite{dwork2006calibrating,abadi2016deep,geyer2017dpfl,mcmahan2018dplm}. This work does not address this attack surface. Instead, we focus exclusively on the threat arising from the data exchange at the cut points, which is unique to SL and for which existing defenses are insufficient in the LLM setting considered here.

\section{Challenges and Key Insights} \label{overview}

\subsection{Why Perturbation-Based Defenses Fail for LLMs}
\label{sec:why_perturb_fail}

Perturbation-based methods are among the most widely adopted defenses in SL for vision models. In that setting, the client often retains only the early portion of the model, while the labels are typically low-dimensional class indices that carry little information on their own. The threat is that the server trains a decoder to reconstruct the client's input from the smashed data produced by the client-side model. Perturbation-based methods counter this by injecting noise into the smashed data to reduce reconstruction quality while preserving task utility~\cite{10.1145/3659154.3659189}.

This defense strategy becomes fundamentally limited in the LLM setting. In LLM fine-tuning, the label sequence is simply the input shifted by one token, so the input and labels are nearly identical. As a result, the server can use its white-box copy of the downstream model to decode the processed smashed data and predict the labels. The difficulty is that this decoder is not an auxiliary attacker model separate from training. It is the same next-token predictor that fine-tuning is trying to make useful. Therefore, perturbing the smashed data strongly enough to make this decoder fail also makes the training path through the LLM fail. Fine-tuning further tightens this tradeoff. Given perturbed smashed data, the client's objective is still to train the model to predict the next token, which also improves the decoder available to the server once the client uploads its updated client-side parameters. The defender therefore faces a difficult tension. Stronger perturbations improve privacy but damage trainability, while weaker perturbations preserve utility but leave label reconstruction effective.

\subsection{The \sysname Approach}
\label{sec:solution}

\begin{figure}[t]
    \centering
    \includegraphics[width=\linewidth]{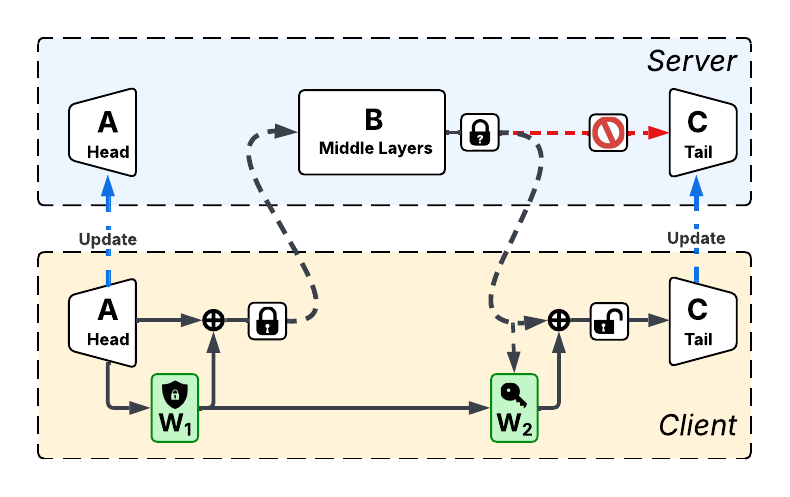}
    \caption{The forward path of \sysname. The client applies an obfuscation adapter $W_1$ to the smashed data before transmission and a recovery adapter $W_2$ to the returned activations. The paired $W_2$ is trained to recover a representation compatible with $C$. Without access to this recovery adapter, applying $f_C$ directly to the obfuscated activations no longer provides an effective decoder.}
    \label{fig:our_model}
\end{figure}

\sysname breaks this single-representation tradeoff by making the server and the client operate on different activation spaces. Rather than adding generic perturbations that must remain usable for training, \sysname applies a learned \textit{obfuscation} to the smashed data before it is transmitted to the server, forcing the server-side computation to proceed in an obfuscated space that is less compatible with the server's decoder. If used alone, this obfuscation would also degrade training because $C$ would receive activations outside its expected distribution. To preserve training effectiveness, \sysname applies a second adapter to the activations returned by the server, mapping them toward a representation compatible with $C$ before the backward pass proceeds on the client side. Conceptually, this forms an obfuscate-and-recover scheme. The client ``\textit{obfuscates}'' the smashed data before sending it to the server, and ``\textit{recovers}'' the returned activations locally, so that the server operates in the obfuscated space while the client's local computation remains aligned with the original model.

The central challenge in realizing this design is the recovery step. Due to the inherent complexity and nonlinearity of LLMs, the obfuscation applied to the smashed data propagates through the server-side model in a highly entangled manner, making it non-trivial to recover a representation compatible with $C$ from the returned activations. The recovery adapter is therefore not assumed to be an exact inverse. Instead, \sysname trains the two adapters jointly. The first adapter learns the forward obfuscation, and the second learns a recovery mapping that preserves training utility. The two adapters are trained simultaneously and cooperate to fulfill the obfuscation and recovery objectives.

\section{Design}
\label{sec:design}

\subsection{Preliminaries}
\label{sec:preliminaries}

We consider an FL system with $N$ clients and one central server. Let $\mathcal{D}_i = \{(x_j, y_j)\}$ denote the private fine-tuning dataset held by client $i$, where $x_j$ is an input token sequence and $y_j$ is the corresponding label sequence. In the LLM fine-tuning setting, $y_j$ is the shifted version of $x_j$, i.e., $y_j^{(t)} = x_j^{(t+1)}$.

Let the LLM consist of an embedding layer followed by $L$ transformer blocks and a language model head. We denote the two cut points as $k$ and $m$ ($1 \le k < m \le L$). The model is then split into three components. $A$ contains the embedding layer and transformer blocks $1$ through $k$. $B$ contains transformer blocks $k{+}1$ through $m$. $C$ contains transformer blocks $m{+}1$ through $L$ and the language model head. The full model satisfies $f = f_C \circ f_B \circ f_A$, and the loss $\mathcal{L}$ is computed at the client. The intermediate activations $s_A = f_A(x)$ and $s_B = f_B(s_A)$, and their corresponding gradients $\nabla_{s_B}\mathcal{L}$ and $\nabla_{s_A}\mathcal{L}$, are exchanged between the client and server at the two cut points.

In each training step, the forward and backward passes follow the split path $f_A \to f_B \to f_C$. The client sends $s_A$ to the server, receives $s_B$, computes the loss through $f_C$, and exchanges the corresponding cut-point gradients with the server during backpropagation. At the end of each round, clients upload their updated client-side parameters or model updates, which the server aggregates via FedAvg before broadcasting the next global model.

The split protocol keeps $B$ frozen and updates only $A$ and $C$. Denoting the parameters of $A$, $B$, and $C$ as $\theta_A$, $\theta_B$, $\theta_C$ respectively, and using learning rate $\eta$ and cross-entropy loss $\ell(\cdot,\cdot)$, the adapter-free updates are given as follows.
\begin{align}
    s_A &= f_A(x),\quad s_B = f_B(s_A) \notag \\[4pt]
    \theta_C &\leftarrow \theta_C - \eta\, \frac{\partial\, \ell\bigl(f_C(s_B),\, y\bigr)}{\partial \theta_C} \\
    \theta_A &\leftarrow \theta_A - \eta\, \Bigl(\frac{\partial f_A}{\partial \theta_A}\Bigr)^\top \Bigl(\frac{\partial f_B}{\partial s_A}\Bigr)^\top \frac{\partial\, \ell\bigl(f_C(s_B),\, y\bigr)}{\partial s_B}
\end{align}

\subsection{Update Protocol and Adapter Design}
\label{sec:protocol}

Recall that $s_A$ is the smashed data produced by the client and transmitted to the server, and $s_B$ is the smashed data returned by the server to the client. \sysname introduces two adapters, $W_1$ and $W_2$, both residing on the client.

\noindent\textbf{Obfuscation adapter $W_1$.} $W_1$ is implemented as a residual Multi-Layer Perceptron (MLP). Letting $h_1$ denote the inner feed-forward network of $W_1$, the obfuscation is computed as follows.
\begin{equation}
    s'_A = W_1(s_A) = s_A + h_1(s_A)
\end{equation}
The obfuscated smashed data $s'_A$ is sent to the server in place of $s_A$. The server passes it through $f_B$ as usual to produce $s'_B = f_B(s'_A)$, and returns $s'_B$ to the client. Note that $W_1$ alone, together with its residual connection, plays the same role as perturbation-based methods, adding a bounded additive perturbation $h_1(s_A)$ to the smashed data before transmission. As with these perturbation-based methods, the intent of $W_1$ is not to fully hide the structure of the smashed data, but rather to add enough noise to deactivate the server's decoder.

\noindent\textbf{Recovery adapter $W_2$.} $W_2$ is implemented as a residual MLP that takes $s'_B$ and the $W_1$ residual $\delta_A = s'_A - s_A$ as inputs. Letting $h_2$ denote the inner feed-forward network of $W_2$, the recovery is computed as follows.
\begin{equation}
    \hat{s}_B = W_2(s'_B,\, \delta_A) = s'_B + h_2\!\left([s'_B \,\|\, \delta_A]\right)
\end{equation}
where $[\cdot \| \cdot]$ denotes concatenation along the feature dimension. The client proceeds with $\hat{s}_B$ in place of $s_B$ for the remainder of the forward and backward passes. Conditioning on $\delta_A$ rather than $s'_A$ directly provides $W_2$ with a compact representation of how $W_1$ modified the activations. This residual assists $W_2$ in recovering a representation compatible with $C$. It is computed from $s_A$ and $s'_A$ on the client and is consumed only as a local input to $W_2$, so it is never transmitted to the server.

\noindent\textbf{STE for the backward pass.}
To decouple adapter behavior from model training, \sysname applies the Straight-Through Estimator (STE)~\cite{bengio2013ste} at both cut points. Concretely, the actual tensors passed through the network are defined as follows.
\begin{align}
    \tilde{s}_A &= s_A + \bigl(s'_A - s_A\bigr)_{\mathrm{sg}} \label{eq:ste_w1} \\
    \tilde{s}_B &= s'_B + \bigl(\hat{s}_B - s'_B\bigr)_{\mathrm{sg}} \label{eq:ste_w2}
\end{align}
where $(\cdot)_{\mathrm{sg}}$ denotes the stop-gradient operator. In the forward pass, $\tilde{s}_A = s'_A$ and $\tilde{s}_B = \hat{s}_B$, so the obfuscated activations and the recovered activations are used as intended. In the backward pass, the stop-gradient term vanishes, so gradients flow through $\tilde{s}_A$ as if it were $s_A$, and through $\tilde{s}_B$ as if it were $s'_B$. STE ensures that $\frac{\partial \mathcal{L}}{\partial s'_B} = \frac{\partial \mathcal{L}}{\partial \tilde{s}_B}$ and $\frac{\partial \tilde{s}_A}{\partial s_A} = I$, so the gradients propagated through $f_B$ and $f_A$ are computed as if $W_1$ and $W_2$ were identity mappings. The STE is specifically designed to keep $A$ trainable, which we discuss further in Section~\ref{sec:trainability}. Note that perturbation-based methods, which directly add noise to the smashed data as $s'_A = s_A + \epsilon$, implicitly apply the same STE treatment in the backward pass.

With the adapters involved, the parameter updates are given as follows.
\begin{align}
    \tilde{s}_A &= f_A(x) + \bigl(W_1(f_A(x)) - f_A(x)\bigr)_{\mathrm{sg}} \notag \\
    \tilde{s}_B &= f_B(\tilde{s}_A) + \bigl(W_2(f_B(\tilde{s}_A),\, \tilde{s}_A - f_A(x)) - f_B(\tilde{s}_A)\bigr)_{\mathrm{sg}} \notag \\[4pt]
    \theta_C &\leftarrow \theta_C - \eta\, \frac{\partial\, \ell\bigl(f_C(\tilde{s}_B),\, y\bigr)}{\partial \theta_C} \\
    \theta_A &\leftarrow \theta_A - \eta\, \Bigl(\frac{\partial f_A}{\partial \theta_A}\Bigr)^\top \Bigl(\frac{\partial f_B}{\partial \tilde{s}_A}\Bigr)^\top \frac{\partial\, \ell\bigl(f_C(\tilde{s}_B),\, y\bigr)}{\partial f_B(\tilde{s}_A)}
\end{align}
$\theta_B$ is kept frozen throughout fine-tuning, and we discuss the reason in Section~\ref{sec:trainability}.

\subsection{Adapter Training}
\label{sec:adapter_training}

$W_1$ serves a dual purpose. Its primary objective is to obfuscate the smashed data into a distribution incompatible with the server's decoder to prevent label leakage, while it must simultaneously preserve enough structure to allow $W_2$ to enable effective training of the main model. $W_2$, on the other hand, is solely dedicated to enabling training. It maps the obfuscated server output back to a representation compatible with $C$. The training losses of $W_1$ and $W_2$ reflect these distinct roles.

Throughout this section, we follow the notation established in Section~\ref{sec:preliminaries}, ignoring STE for now. $s_A$ and $s_B$ denote the smashed data produced without any adapter obfuscation. $s'_A$ and $s'_B$ denote the activations after $W_1$ has been applied but before $W_2$, i.e., $s'_A = W_1(s_A)$ and $s'_B = f_B(s'_A)$. $\hat{s}_B$ denotes the final output after both adapters have been applied, i.e., $\hat{s}_B = W_2(s'_B,\, \delta_A)$ where $\delta_A = s'_A - s_A$. We further denote by $z = f_C(s_B)$, $z' = f_C(s'_B)$, and $\hat{z} = f_C(\hat{s}_B)$ the logits produced by $f_C$ under the plain, $W_1$-only, and full-adapter paths, respectively. 

\subsubsection{Loss of \texorpdfstring{$W_1$}{W1}}
Let $P$ denote the set of valid token positions, $V$ the vocabulary size, and $\tau$ a temperature hyperparameter. For a logit tensor $z$, define the token-level softmax distribution $p^z_t = \mathrm{softmax}(z_t / \tau) \in \mathbb{R}^V$, and let $m_t(z,z') = \frac{1}{2}(p^z_t + p^{z'}_t)$. The token-averaged Jensen-Shannon divergence is given as follows.
\begin{align}
    D_{\mathrm{JSD}}(z, z') &= \frac{\tau^2}{2|P|}\sum_{t \in P}\sum_{v=1}^{V}\Bigl(p^z_{t,v}\log\frac{p^z_{t,v}}{m_{t,v}(z,z')} \notag \\
    &\qquad + p^{z'}_{t,v}\log\frac{p^{z'}_{t,v}}{m_{t,v}(z,z')}\Bigr)
\end{align}
For the CKA term, let $\tilde{X}$ denote the column-mean-centered token matrix of $X$ restricted to $P$. The linear CKA is defined as follows.
\begin{equation}
    \mathrm{CKA}(X, Y) = \frac{\|\tilde{X}^\top \tilde{Y}\|_F^2}{\|\tilde{X}^\top \tilde{X}\|_F \cdot \|\tilde{Y}^\top \tilde{Y}\|_F}
\end{equation}
The training objective of $W_1$ is then defined as follows.
\begin{align}
    \mathcal{L}_{W_1} &= \lambda_{\text{CKA}}^{(1)}\, \mathrm{CKA}(s'_B,\, s_B) \label{eq:w1_cka} \\
    &- \lambda_{\text{JSD}}^{(1)}\, D_{\mathrm{JSD}}(z',\, z) \label{eq:w1_ntkl} \\
    &+ \lambda_{W_2}^{(1)}\, \mathcal{L}_{W_2} \label{eq:w1_w2}
\end{align}

The term \eqref{eq:w1_ntkl} is the most direct privacy objective. Maximizing the JSD between $z'$ and $z$ reduces the server's ability to correctly decode the transmitted smashed data $s'_A$ using $f_C \circ f_B$. The CKA term \eqref{eq:w1_cka} provides a further and complementary signal that operates at the representation level rather than the logit level, as minimizing it pushes the geometry of $s'_B$ away from $s_B$, making the intermediate activations themselves harder for the server to exploit even before reaching $C$. Together, the two terms reduce decodability at two different levels of the server-side model.

The term \eqref{eq:w1_w2} couples $W_1$ to the restoration objective of $W_2$. While the privacy terms push $W_1$ to produce an obfuscation that is hard to decode, \eqref{eq:w1_w2} simultaneously encourages $W_1$ to produce obfuscations for which $W_2$ can keep the reconstruction error small. This coupling is essential. Without it, $W_1$ may produce transformations that are trivially obfuscated but too difficult for $W_2$ to map back to a trainable representation, collapsing training effectiveness. $W_1$ and $W_2$ are therefore mutually dependent and must be trained jointly. The weight $\lambda_{W_2}^{(1)}$ controls the strength of this coupling relative to the privacy terms \eqref{eq:w1_cka} and \eqref{eq:w1_ntkl}. A larger $\lambda_{W_2}^{(1)}$ favors utility by biasing $W_1$ toward obfuscations that $W_2$ can more easily map back to a trainable representation, while a smaller value favors privacy by allowing $W_1$ to pursue stronger obfuscation at the risk of degrading $W_2$'s recovery quality.

\subsubsection{Loss of \texorpdfstring{$W_2$}{W2}}
Recall that $\hat{z}$ denotes the logits produced via the full adapter path ($W_1$ followed by $W_2$), and $\hat{s}_B = W_2(s'_B,\, \delta_A)$ denotes the reconstructed smashed data. Following the notation above, the token-averaged KL divergence is given as follows.
\begin{equation}
    D_{\mathrm{KL}}(z \| z') = \frac{\tau^2}{|P|} \sum_{t \in P} \sum_{v=1}^{V} p^z_{t,v} \log \frac{p^z_{t,v}}{p^{z'}_{t,v}}
\end{equation}
The training objective of $W_2$ is then defined as follows.
\begin{align}
    \mathcal{L}_{W_2} &= \lambda_{\text{KL}}^{(2)}\, D_{\mathrm{KL}}\!\bigl(\hat{z} \,\|\, z\bigr) \label{eq:w2_kl} \\
    &+ \lambda_{\text{MSE}}^{(2)}\, \bigl\|\hat{s}_B - s_B\bigr\|^2 \label{eq:w2_mse}
\end{align}

The term \eqref{eq:w2_mse} directly supervises $W_2$ to reconstruct the original smashed data $s_B$ from the obfuscated server output $s'_B$. However, due to the inherent complexity and nonlinearity of the LLM, a perfect reconstruction is in general unattainable. The term \eqref{eq:w2_kl} therefore provides a complementary signal. Even when $\hat{s}_B$ cannot exactly match $s_B$, minimizing $D_{\mathrm{KL}}(\hat{z} \| z)$ encourages the output distribution of $f_C$ to match the plain-path distribution, compensating for residual reconstruction error at the distribution level. Together, the two terms help keep $C$ trainable despite the imperfect recovery from $W_1$'s obfuscation.

\subsection{Trainability of the LLM Model}
\label{sec:trainability}

We now give an intuitive analysis of the trainability of $A$, $B$, and $C$ under the adapter protocol. Formal analysis is provided in Section~\ref{sec:trainability_analysis}.

\noindent\textbf{Trainability of $C$.}
Suppose $W_2$ is perfectly trained, i.e., $\hat{s}_B = s_B$ exactly. Then from the perspective of $C$, the presence of the adapters is entirely transparent. It receives the same input as in the adapter-free setting, and its training proceeds identically. Although $\hat{s}_B = s_B$ is unattainable in practice, as long as the residual error $\|\hat{s}_B - s_B\|$ is small, it can be treated as noise and the training effectiveness of $C$ is preserved.

\noindent\textbf{Trainability of $B$.}
We keep $B$ frozen throughout training. Under the adapter protocol, $B$ receives $s'_A = W_1(s_A)$ as input rather than $s_A$. Because $W_1$ deliberately pushes $s'_A$ out of the distribution that $B$ was originally trained on, any updates to $\theta_B$ would reflect the obfuscated space induced by $W_1$, not the original input distribution. Deploying the fine-tuned $B$ without the adapters would expose it to out-of-distribution inputs, rendering the fine-tuning ineffective. Retaining $W_1$ during inference would require the client to apply it on every query, which conflicts with the standard FL deployment goal of a complete, independently usable model on the server. For fine-tuning purposes, updating $A$ and $C$ alone is sufficient to adapt the model to downstream tasks, as confirmed by our experiments in Section~\ref{sec:exp_training}.

\noindent\textbf{Trainability of $A$.}
The adapter protocol introduces a gradient deviation in $A$ proportional to $\|s'_A - s_A\|$. If $W_2$ were absent, \sysname would degenerate into a perturbation-based scheme, and trainability would follow the same condition as prior work. The introduction of $W_2$ relaxes this constraint when the recovery error $\|\hat{s}_B - s_B\|$ remains small, because $C$ then receives a representation closer to the plain-path smashed data even when $s'_A$ deviates from $s_A$. We provide an analysis in Section~\ref{sec:comparison_perturb}.

\subsection{Full Training Protocol}
\label{sec:full_protocol}

\noindent\textbf{Alignment dataset.}
Training $W_1$ and $W_2$ uses an arbitrary alignment dataset $\mathcal{D}_{\mathrm{align}} = \{(x_j, y_j)\}_{j=1}^{M}$ that contains no private information. This dataset is independent of the private fine-tuning dataset $\mathcal{D}_i$ and does not need to match its task, domain, or distribution. It may come from any public or synthetic text source that can be processed by the underlying model.

\noindent\textbf{Adapter alignment procedure.}
Given $\mathcal{D}_{\mathrm{align}}$, the client runs a forward pass through the full adapter path $f_A \to W_1 \to f_B \to W_2 \to f_C$ as well as the plain path $f_A \to f_B \to f_C$. For each batch, the client computes the intermediate activations $s_A,\, s'_A,\, s_B,\, s'_B,\, \hat{s}_B$ and the corresponding logits $z,\, z',\, \hat{z}$ as defined above. The adapters are then updated using their objectives. $W_1$ is updated by minimizing $\mathcal{L}_{W_1}$ defined in \eqref{eq:w1_cka} through \eqref{eq:w1_w2}, and $W_2$ is updated by minimizing $\mathcal{L}_{W_2}$ defined in \eqref{eq:w2_kl} through \eqref{eq:w2_mse}. All other model parameters, including $\theta_A$ and $\theta_C$, remain frozen throughout alignment, so the alignment procedure only ever updates $W_1$ and $W_2$. The alignment phase runs for a fixed number of steps prior to the main fine-tuning loop.

\noindent\textbf{Main fine-tuning loop.}
After the adapter alignment phase, the system proceeds with the standard FL fine-tuning loop. At each global round $r$, the server selects a subset of clients $\mathcal{S}^{(r)} \subseteq \{1, \ldots, N\}$ and broadcasts the current model parameters $\theta_A^{(r)},\, \theta_C^{(r)}$ to each selected client. Each client $i \in \mathcal{S}^{(r)}$ then performs $E$ local steps on its private dataset $\mathcal{D}_i$. In each local step, the client runs the full adapter forward pass to obtain $\hat{z}$, computes the cross-entropy loss $\mathcal{L} = \ell(\hat{z}, y_{\mathrm{gt}})$ against the ground-truth labels $y_{\mathrm{gt}}$, and updates $\theta_A$ and $\theta_C$ via the STE-based backward pass described above, with $\theta_B$ kept frozen. After $E$ local steps, each client uploads the updated $\theta_A^{(r,i)}$ and $\theta_C^{(r,i)}$ to the server. The server then aggregates the updates via FedAvg, as follows.
\begin{equation}
    \theta^{(r+1)} \leftarrow \sum_{i \in \mathcal{S}^{(r)}} \frac{|\mathcal{D}_i|}{\sum_{j \in \mathcal{S}^{(r)}} |\mathcal{D}_j|}\, \theta^{(r,i)}
\end{equation}
where $\theta$ denotes $\theta_A$ and $\theta_C$ jointly. The adapters $W_1$ and $W_2$ are never uploaded and remain local to each client throughout. In the general setting, each client independently trains and maintains its own adapter pair $(W_1^{(i)}, W_2^{(i)})$. Among mutually trusted clients, adapters may be shared when their alignment data policy permits sharing, as illustrated in Figure~\ref{fig:fl_plane}.

Because $W_1$ and $W_2$ are trained against a specific snapshot of $\theta_A$ and $\theta_C$, they may become misaligned as the global model $\theta$ continues to accumulate updates over successive rounds. To address this, after a period of training steps, each client enters a short burst alignment phase in which $W_1$ and $W_2$ are re-trained on $\mathcal{D}_{\mathrm{align}}$ for a few steps against the current model weights to recalibrate the adapter pair before the next training phase resumes.

Once the fine-tuning loop concludes, the server holds the complete fine-tuned model $f = f_C \circ f_B \circ f_A$, which is ready for deployment. The complete protocol is summarized in Algorithm~\ref{alg:protocol}.

\begin{algorithm}[t]
\caption{\sysname Full Training Protocol}
\label{alg:protocol}
\begin{algorithmic}[1]
\Require Private dataset $\mathcal{D}_i$, alignment dataset $\mathcal{D}_{\mathrm{align}}$, initial parameters $\theta_A^{(0)}, \theta_B, \theta_C^{(0)}$, global rounds $R$, local steps $E$, alignment steps $T_{\mathrm{align}}$, burst interval $K$
\Ensure Fine-tuned model $\theta_A^{(R)}, \theta_B, \theta_C^{(R)}$
\State \textbf{// Initial adapter alignment}
\State Run alignment on $\mathcal{D}_{\mathrm{align}}$ for $T_{\mathrm{align}}$ steps with $\theta_A, \theta_C$ frozen, and update $W_1, W_2$ with $\mathcal{L}_{W_1}, \mathcal{L}_{W_2}$
\For{each global round $r = 1, \ldots, R$}
    \State Server broadcasts $\theta_A^{(r)}, \theta_C^{(r)}$ to selected clients $\mathcal{S}^{(r)}$
    \For{each client $i \in \mathcal{S}^{(r)}$ \textbf{in parallel}}
        \For{$e = 1, \ldots, E$ local steps}
            \State Sample batch from $\mathcal{D}_i$
            \State Forward: $s_A \to W_1 \to s'_A \to f_B \to s'_B \to W_2 \to \hat{s}_B \to f_C \to \hat{z}$
            \State Compute loss $\mathcal{L} = \ell(\hat{z}, y_{\mathrm{gt}})$
            \State Backward via STE; update $\theta_A, \theta_C$ (freeze $\theta_B$)
        \EndFor
        \If{$r \bmod K = 0$} \textbf{// Burst alignment}
            \State Re-train $W_1, W_2$ on $\mathcal{D}_{\mathrm{align}}$ for a few steps against current $\theta$
        \EndIf
        \State Upload $\theta_A^{(r,i)}, \theta_C^{(r,i)}$ to server
    \EndFor
    \State Server aggregates: $\theta^{(r+1)} \leftarrow \mathrm{FedAvg}\bigl(\{\theta^{(r,i)}\}_{i \in \mathcal{S}^{(r)}}\bigr)$
\EndFor
\end{algorithmic}
\end{algorithm}

\section{Trainability \& Convergence Analysis}
\label{sec:analysis}
\subsection{Trainability Analysis}
\label{sec:trainability_analysis}

We begin with the single-client setting, abstracting away FL and data heterogeneity. We formalize the conditions under which the adapter protocol yields an expected decrease in the plain-path loss. Let $\mathcal{L}$ denote the plain-path loss and $g_\theta^{\mathrm{plain}} = \partial \mathcal{L} / \partial \theta$ the ideal gradient. The adapter-path gradient actually used for updates is denoted $g_\theta^{\mathrm{adapter}}$. We assume $\mathcal{L}$ is $L_F$-smooth in $\theta$, in which case a step $\theta \leftarrow \theta - \eta g_\theta^{\mathrm{adapter}}$ with sufficiently small $\eta$ decreases $\mathcal{L}$ whenever $\langle g_\theta^{\mathrm{adapter}}, g_\theta^{\mathrm{plain}} \rangle > 0$. A sufficient condition is
\begin{equation} \label{eq:descent_cond}
    \lVert \delta g_\theta \rVert < \lVert g_\theta^{\mathrm{plain}} \rVert, \quad \delta g_\theta := g_\theta^{\mathrm{adapter}} - g_\theta^{\mathrm{plain}}.
\end{equation}
We make the following smoothness assumptions on $f_B$'s Jacobian and on $\ell \circ f_C$.

\begin{assumption} \label{as:jb}
\textit{(i) The Jacobian $J_B$ of $f_B$ is $L_{J_B}$-Lipschitz: $\lVert J_B(u) - J_B(v) \rVert_{\mathrm{op}} \leq L_{J_B} \lVert u - v \rVert$ for all $u, v$, and $f_B$ itself is $L_B$-Lipschitz with $\lVert J_B(u) \rVert_{\mathrm{op}} \leq L_B$. (ii) $\ell \circ f_C$ is $L_C$-smooth in its first argument $s$, and the cross derivative $\partial \nabla_{\theta_C}(\ell \circ f_C)/\partial s$ is also bounded by $L_C$ in operator norm. (iii) The upstream gradient $g_{\mathrm{up}} = \partial \mathcal{L}/\partial s_B$ and the parameter Jacobians $\partial f_A/\partial \theta_A$, $\partial f_C/\partial \theta_C$ have bounded operator norms uniformly along the optimization trajectory.}
\end{assumption}

Such Lipschitz-smoothness assumptions are standard throughout non-convex optimization for justifying descent-based arguments, and here they are applied to $f_B$'s Jacobian and to $\ell \circ f_C$. If they fail to hold, gradient descent itself carries no guarantee of decreasing the loss, regardless of whether \sysname's adapters are used.

Let $J_{W_1}$ denote the Jacobian of $W_1$, and let $J_{W_2}^{(1)}$, $J_{W_2}^{(2)}$ denote the partial Jacobians of $W_2$ with respect to its first input ($s'_B$) and second input ($\delta_A$). Note that $s_A$ enters the adapter chain through two paths: indirectly via $s'_A = W_1(s_A)$ (which feeds both $f_B$ and $\delta_A$), and directly via the $-s_A$ term in $\delta_A = s'_A - s_A$. Before applying STE, direct backpropagation through the adapter chain $f_A \to W_1 \to f_B \to W_2 \to f_C$ gives
\begin{align}
    g_{\theta_C}^{\mathrm{adapter}} &= \frac{\partial\, \ell(f_C(\hat{s}_B), y)}{\partial \theta_C}, \label{eq:gc_full} \\
    g_{\theta_A}^{\mathrm{adapter}} &= \Bigl(\frac{\partial f_A}{\partial \theta_A}\Bigr)^{\!\top}\!\Bigl[J_{W_1}(s_A)^{\!\top} J_B(s'_A)^{\!\top} J_{W_2}^{(1)}(s'_B, \delta_A)^{\!\top} \notag \\
    &\quad + \bigl(J_{W_1}(s_A)^{\!\top} - I\bigr) J_{W_2}^{(2)}(s'_B, \delta_A)^{\!\top}\Bigr] \frac{\partial \mathcal{L}}{\partial \hat{s}_B}. \label{eq:ga_full}
\end{align}
The corresponding plain-path gradients are $$g_{\theta_C}^{\mathrm{plain}} = \partial\, \ell(f_C(s_B), y) / \partial \theta_C \text{ and}$$  $$g_{\theta_A}^{\mathrm{plain}} = (\partial f_A / \partial \theta_A)^{\!\top} J_B(s_A)^{\!\top} \partial \mathcal{L} / \partial s_B.$$

Comparing \eqref{eq:ga_full} against the plain-path expression, the deviation $\delta g_{\theta_A}$ contains four sources, namely (i)~$J_{W_1}$ deviating from identity, (ii)~$J_{W_2}^{(1)}$ deviating from identity, (iii)~$J_{W_2}^{(2)}$ deviating from zero, and (iv)~the mismatch between $\partial \mathcal{L} / \partial \hat{s}_B$, $J_B(s'_A)$ and $\partial \mathcal{L} / \partial s_B$, $J_B(s_A)$. The coupling term $(J_{W_1}^{\!\top} - I)\, J_{W_2}^{(2)\top}$ in \eqref{eq:ga_full} vanishes whenever either (i) or (iii) is suppressed.

Under STE, sources (i)--(iii) vanish. The stop-gradient construction of Section~\ref{sec:design} enforces
\begin{equation}
    J_{W_1}(s_A) = I, \quad J_{W_2}^{(1)}(s'_B, \delta_A) = I, \quad J_{W_2}^{(2)}(s'_B, \delta_A) = 0
\end{equation}
in the backward pass, independent of the actual functional form of $W_1$ and $W_2$. When substituting into \eqref{eq:ga_full}, only source (iv) remains, and it decomposes into an upstream gradient error and a Jacobian evaluation error:
\begin{align}
    \delta g_{\theta_A} &= \Bigl(\frac{\partial f_A}{\partial \theta_A}\Bigr)^{\!\top}\!\Bigl[J_B(s'_A)^{\!\top}\!\Bigl(\tfrac{\partial \mathcal{L}}{\partial \hat{s}_B} - \tfrac{\partial \mathcal{L}}{\partial s_B}\Bigr) \notag \\
    &\quad + \bigl(J_B(s'_A) - J_B(s_A)\bigr)^{\!\top}\!\tfrac{\partial \mathcal{L}}{\partial s_B}\Bigr]. \label{eq:delta_g_decomp}
\end{align}

Using Assumption~\ref{as:jb} on \eqref{eq:delta_g_decomp}, and letting $g_{\mathrm{up}} = \partial \mathcal{L} / \partial s_B$, we obtain
\begin{align}
    \lVert \delta g_{\theta_A} \rVert &\leq \Bigl\lVert \tfrac{\partial f_A}{\partial \theta_A} \Bigr\rVert_{\mathrm{op}} \Bigl[\lVert J_B(s'_A) \rVert_{\mathrm{op}} \cdot L_C \lVert \hat{s}_B - s_B \rVert \notag \\
    &\quad + \lVert g_{\mathrm{up}} \rVert \cdot L_{J_B} \lVert s'_A - s_A \rVert\Bigr]. \label{eq:delta_g_bound}
\end{align}
An analogous bound for $\theta_C$ gives $\lVert \delta g_{\theta_C} \rVert \leq L_C \lVert \hat{s}_B - s_B \rVert$.

Combining \eqref{eq:descent_cond} and \eqref{eq:delta_g_bound} yields the sufficient condition for $\theta_A$:
\begin{align} \label{eq:train_A_condition}
    \lVert J_B(s'_A) \rVert_{\mathrm{op}} \cdot& L_C \lVert \hat{s}_B - s_B \rVert + \lVert g_{\mathrm{up}} \rVert \cdot L_{J_B} \lVert s'_A - s_A \rVert \notag\\ 
    & \quad  < \frac{\lVert g_{\theta_A}^{\mathrm{plain}} \rVert}{\lVert \partial f_A / \partial \theta_A \rVert_{\mathrm{op}}},
\end{align}
and analogously $L_C \lVert \hat{s}_B - s_B \rVert < \lVert g_{\theta_C}^{\mathrm{plain}} \rVert$ for $\theta_C$. The first error term in \eqref{eq:train_A_condition} is directly minimized by the $W_2$ loss \eqref{eq:w2_mse}, while the second is kept finite by the residual structure of $W_1$ together with the coupling term \eqref{eq:w1_w2} described in Section~\ref{sec:protocol}.

\subsection{Convergence Analysis}
\label{sec:convergence_analysis}
We now extend to the full FL setting with $N$ clients and non-i.i.d.\ data. Our convergence analysis builds on the standard FedAvg convergence framework of~\cite{Li2020On}, adopting Assumptions~\ref{as:1} through \ref{as:5} below to characterize the underlying FL optimization problem, and introduces an additional smashed data deviation constraint, Assumption~\ref{as:6}. If any of Assumptions~\ref{as:1} through \ref{as:5} fails to hold, standard FL itself is not guaranteed to converge, regardless of whether \sysname is used. We first analyze the case where $\theta_A$ is frozen and only $\theta_C$ is trained. We extend to joint $(\theta_A, \theta_C)$ training at the end of this section. Throughout, $\theta$ refers to $\theta_C$ unless stated otherwise, and $\theta_B$ is frozen throughout (see Section~\ref{sec:design}). Let $F_i(\theta)$ denote the local objective at client $i$ under the plain (adapter-free) path.

\begin{assumption} \label{as:1}
\textit{Local objective functions} $F_1, \dots, F_N$ \textit{are all L-smooth: for all v and w,} $F_i(v) \leq F_i(w) + (v-w)^T\nabla F_i(w) + \frac{L}{2}\left\lVert v-w \right\rVert^2_2$.
\end{assumption}

\begin{assumption} \label{as:2}
$F_1, \dots, F_N$ \textit{are all $\mu$-strongly convex: for all v and w,} $F_i(v) \geq F_i(w) + (v-w)^T\nabla F_i(w) + \frac{\mu}{2}\left\lVert v-w \right\rVert^2_2$.
\end{assumption}

We adopt Assumption~\ref{as:2}, together with Assumptions~\ref{as:1}, \ref{as:3}--\ref{as:5}, from the FedAvg analysis of~\cite{Li2020On} to characterize the underlying FL optimization problem in a setting where a convergence guarantee is analytically tractable, following standard practice in the FL convergence literature. Like this and other such analyses, it does not literally hold for the non-convex loss landscape of LLM fine-tuning. Within this same idealized setting, our point is that \sysname introduces no convergence obstacle beyond what the underlying FedAvg baseline already requires.

\begin{assumption} \label{as:3}
\textit{Let $\xi_i$ be sampled from the i-th device's local data uniformly at random. The variance of stochastic gradients in each device is bounded: $\mathbb{E}\left\lVert \nabla F_i(\xi_i;\theta_i^t) - \nabla F_i(\theta_i^t)\right\rVert^2 \leq \sigma^2_i$ for $i=1,\dots,N$.}
\end{assumption}

\begin{assumption} \label{as:4}
\textit{The expected squared norm of stochastic gradients is uniformly bounded, i.e., $\mathbb{E}\left\lVert \nabla F_i(\xi_i;\theta_i^t)\right\rVert^2 \leq G^2$ for all $i=1,\dots,N$ and $t=1,\dots,T-1$.}
\end{assumption}

\begin{assumption} \label{as:5}
\textit{Assume $S_t$ contains a subset of K indices uniformly randomly from [N] without replacement. Assume the data is balanced in the sense that $p_1 = \dots = p_N = \frac{1}{N}$. The aggregation step of FedAvg performs $\theta_t \leftarrow \frac{N}{K}\sum_{i \in S_t} p_i\theta_i^t$}.
\end{assumption}

\begin{assumption} \label{as:6}
\textit{The expected squared norm of the difference between the plain-path smashed data $s_B$ and the recovered smashed data $\hat{s}_B$ is bounded. Specifically, $\mathbb{E}\left\lVert s_B - \hat{s}_B \right\rVert^2 \leq H$ for all $i=1,\dots,N$ and $t=1,\dots,T-1$.}
\end{assumption}

By Assumption~\ref{as:jb} and the bound \eqref{eq:delta_g_bound}, there exists a constant $L_s > 0$ depending only on the quantities in Assumption~\ref{as:jb} such that, in the $\theta_C$-only setting,
\begin{equation} \label{eq:Ls_bound}
    \mathbb{E}\bigl\lVert \tilde{\nabla} F_i(\xi_i;\theta) - \nabla F_i(\xi_i;\theta) \bigr\rVert^2 \leq L_s^2\, \mathbb{E}\bigl\lVert s_B - \hat{s}_B \bigr\rVert^2,
\end{equation}
where $\tilde{\nabla} F_i$ denotes the adapter-path gradient and $\nabla F_i$ denotes the plain-path gradient at the same parameters. For $\theta_C$, $L_s = L_C$ follows directly from the cross-derivative bound in Assumption~\ref{as:jb}.

Now we derive a variance bound that accounts for the adapter-path gradient in place of Assumption~\ref{as:3}. Let $\delta^t_i = \tilde{\nabla} F_i(\xi_i;\theta_i^t)$. By the elementary inequality $\lVert a + b \rVert^2 \leq 2\lVert a \rVert^2 + 2\lVert b \rVert^2$ together with \eqref{eq:Ls_bound} and Assumptions~\ref{as:3},~\ref{as:6},
\begin{align}
    & \mathbb{E}\left\lVert \delta^t_i - \nabla F_i(\theta_i^t)\right\rVert^2 \notag \\
    &\leq 2\,\mathbb{E}\left\lVert \delta^t_i - \nabla F_i(\xi_i;\theta_i^t)\right\rVert^2 + 2\,\mathbb{E}\left\lVert \nabla F_i(\xi_i;\theta_i^t) - \nabla F_i(\theta_i^t)\right\rVert^2 \notag \\
    &\leq 2 L_s^2\, \mathbb{E}\left\lVert s_B - \hat{s}_B \right\rVert^2 + 2\sigma^2_i \notag \\
    &\leq 2(\sigma^2_i + L_s^2 H).
\end{align}
Note that this bound subsumes both the variance and the bias of $\delta^t_i$ relative to $\nabla F_i(\theta_i^t)$, since $\mathbb{E}\lVert \delta - \nabla F \rVert^2 = \lVert \mathbb{E}\delta - \nabla F \rVert^2 + \mathrm{Var}(\delta)$. The same argument yields a corresponding squared-norm bound in place of Assumption~\ref{as:4}, namely $\mathbb{E}\left\lVert \delta^t_i \right\rVert^2 \leq 2(G^2 + L_s^2 H)$ for all $i=1,\dots,N$ and $t=1,\dots,T-1$.

We define $F^*$ and $F^*_i$ as the minimum values of $F$ and $F_i$ and $\Gamma = F^* - \sum^N_{i=1} p_i F^*_i$. We assume each device has $E$ local updates. The bound above shows that the adapter-path update can be viewed as a stochastic update whose deviation from the plain-path gradient is controlled by the recovery error $H$. Applying the FedAvg convergence argument of~\cite{Li2020On} with the gradient-deviation bound $2(\sigma^2_i + L_s^2 H)$ and the squared-norm bound $2(G^2 + L_s^2 H)$ gives the following bound on non-i.i.d.\ data.

\begin{theorem}

\textit{
Let Assumption \ref{as:jb} and Assumptions~\ref{as:1}--\ref{as:6} hold, and let $L,\mu,\sigma_i, G, K, H, L_s$ be defined therein. Choose $\kappa = \frac{L}{\mu}, \gamma = \max\{8\kappa, E\}$ and the learning rate $\eta_t=\frac{2}{\mu(\gamma + t)}$. Then}\\
\begin{equation}
    \mathbb{E}[F(\theta_T)] - F^* \leq \frac{\kappa}{\gamma + T - 1}\!\left(\frac{2(P+Q)}{\mu} + \frac{\mu\gamma}{2}\mathbb{E}\left\lVert \theta_1 - \theta^* \right\rVert^2\right)
\end{equation}
\textit{where}
\begin{equation}
    P = 2\!\sum^N_{i=1} p^2_i(\sigma^2_i + L_s^2 H ) + 6L\Gamma + 16(E-1)^2(G^2 + L_s^2 H ),
\end{equation}
\begin{equation}
    Q = \frac{N-K}{N-1}\frac{8}{K}E^2(G^2+L_s^2 H ).
\end{equation}
\end{theorem}

\noindent\textbf{Extension to joint $(\theta_A, \theta_C)$ training.}
When $\theta_A$ is also trained, the adapter-path gradient deviation for $\theta_A$ contains an additional term proportional to $\lVert s'_A - s_A \rVert$ (cf.\ \eqref{eq:delta_g_bound}), which is not captured by Assumption~\ref{as:6}. To handle this, we introduce one further assumption.

\begin{assumption} \label{as:7}
\textit{The expected squared norm of the difference between the plain-path smashed data $s_A$ and the obfuscated smashed data $s'_A$ is bounded. Specifically, $\mathbb{E}\left\lVert s_A - s'_A \right\rVert^2 \leq H_A$ for all $i=1,\dots,N$ and $t=1,\dots,T-1$.}
\end{assumption}

Under Assumption~\ref{as:jb} and Assumptions~\ref{as:1}--\ref{as:7}, applying the bound \eqref{eq:delta_g_bound} together with $(a+b)^2 \leq 2a^2 + 2b^2$ shows that there exists $L_s > 0$ such that
\begin{equation}
    \mathbb{E}\bigl\lVert \tilde{\nabla} F_i(\xi_i;\theta) - \nabla F_i(\xi_i;\theta) \bigr\rVert^2 \leq L_s^2\,(H + H_A).
\end{equation}
The derivation then proceeds as before, with the gradient variance bound replaced by $2\bigl(\sigma^2_i + L_s^2(H + H_A)\bigr)$ and the squared gradient norm bound replaced by $2\bigl(G^2 + L_s^2(H + H_A)\bigr)$. The theorem above then holds with $P$ and $Q$ replaced by
\begin{align}
    P &= 2\!\sum^N_{i=1} p^2_i\bigl(\sigma^2_i + L_s^2(H+H_A)\bigr) + 6L\Gamma \notag \\
    &\quad + 16(E-1)^2\bigl(G^2 + L_s^2(H+H_A)\bigr), \\
    Q &= \frac{N-K}{N-1}\frac{8}{K}E^2\bigl(G^2+L_s^2(H+H_A)\bigr).
\end{align}

\subsection{Comparison with Perturbation-Based Defenses}
\label{sec:comparison_perturb}

In a perturbation-based scheme, $s'_A = s_A + \varepsilon$, and the client feeds $s'_B = f_B(s'_A)$ directly to $C$. In this case, the upstream gradient error in \eqref{eq:delta_g_bound} is $L_C\lVert s'_B - s_B\rVert$, and the convergence bound contains $H_{\mathrm{perturb}} = \mathbb{E}\lVert s'_B - s_B\rVert^2$ in place of $H$. By Assumption~\ref{as:jb}, $f_B$ is $L_B$-Lipschitz, so
\begin{equation}
    H_{\mathrm{perturb}} = \mathbb{E}\lVert s'_B - s_B \rVert^2 \leq L_B^2\, H_A,
\end{equation}
That is, $H_{\mathrm{perturb}}$ is \emph{implicitly} controlled by $H_A$. The two are coupled through $L_B$, and there is only one degree of freedom. Consequently, in any perturbation-based scheme the convergence bound contains $L_s^2(H_{\mathrm{perturb}} + H_A) \leq L_s^2(L_B^2 + 1) H_A$, and both error terms in \eqref{eq:train_A_condition} also grow together as the perturbation magnitude $\lVert\varepsilon\rVert$ increases. Strengthening the privacy protection through a larger $\varepsilon$ thus directly worsens the convergence-rate constants $P, Q$ and tightens the descent margin in \eqref{eq:train_A_condition}.

In \sysname, $H = \mathbb{E}\lVert \hat{s}_B - s_B\rVert^2$ is \emph{explicitly} controlled by the $W_2$ loss \eqref{eq:w2_mse}, while $H_A = \mathbb{E}\lVert s'_A - s_A\rVert^2$ is a separate quantity shaped by the residual structure of $W_1$ together with the coupling term \eqref{eq:w1_w2}. Thus the training quality depends on the achieved pair $(H,H_A)$ rather than on $H_A$ alone. When the recovery adapter attains a small value of $H$, a larger deviation at the first cut point can be tolerated without proportionally increasing the error seen by $C$. The advantage of \sysname is therefore captured by Assumptions~\ref{as:6} and~\ref{as:7}. \sysname decouples $H_A$ from $H$, unlike the perturbation-based case where the two are tied together through $L_B$, so privacy can be strengthened as long as $W_2$ keeps the corresponding reconstruction error $H$ small.

\section{Evaluation}
\label{sec:experiments}

\begin{table*}[!ht]
  \centering
  \caption{Defense capability comparison. Each cell reports the mean ROUGE-1 of reconstructed private inputs under the corresponding attack, measured at the final checkpoint. Higher ROUGE-1 indicates stronger leakage, and lower is better for the defense. DF denotes Direct Forward. Boldface marks the best defense in each column.}
  \label{tab:defense}
  \footnotesize
  \setlength{\tabcolsep}{2.4pt}
  \resizebox{\textwidth}{!}{%
  \begin{tabular}{llcccccccccccc}
    \toprule
    & & \multicolumn{4}{c}{\textbf{Banking77}} & \multicolumn{4}{c}{\textbf{CLINC150}} & \multicolumn{4}{c}{\textbf{MentalChat16K}} \\
    \cmidrule(lr){3-6} \cmidrule(lr){7-10} \cmidrule(lr){11-14}
    \textbf{Model} & \textbf{Method} & \textbf{DF} & \textbf{TAG} & \textbf{LAMP} & \textbf{BiSR} & \textbf{DF} & \textbf{TAG} & \textbf{LAMP} & \textbf{BiSR} & \textbf{DF} & \textbf{TAG} & \textbf{LAMP} & \textbf{BiSR} \\
    \midrule
    \multirow{5}{*}{Llama-3.2-3B}
      & No Defense                      & 0.1223 & 0.5336 & 0.7728 & 0.9991 & 0.0644 & 0.8001 & 0.8103 & 1.0000 & 0.6488 & 0.9900 & 0.9927 & 0.9916 \\
      & Embedding $\varepsilon$-Privacy & 0.0572 & 0.6447 & 0.7119 & 0.8181 & 0.0266 & 0.6506 & 0.5524 & 0.8140 & 0.1895 & 0.9971 & 0.9969 & 0.9876 \\
      & Smashed-Data DP                 & 0.1095 & 0.5695 & 0.7749 & 0.9991 & 0.0532 & 0.7305 & 0.7973 & 0.9994 & 0.3337 & 0.9970 & 0.9960 & 0.9956 \\
      & NoPeek                          & 0.1183 & 0.5547 & 0.8573 & 0.9428 & 0.0740 & 0.7602 & 0.8595 & 1.0000 & 0.5453 & 0.9954 & 0.9938 & 0.9998 \\
      & \sysname (ours)                 & \textbf{0.0000} & \textbf{0.0000} & \textbf{0.0000} & \textbf{0.0000} & \textbf{0.0000} & \textbf{0.0000} & \textbf{0.0000} & \textbf{0.0680} & \textbf{0.0000} & \textbf{0.0000} & \textbf{0.0000} & \textbf{0.1126} \\
    \midrule
    \multirow{5}{*}{Llama-3.1-8B}
      & No Defense                      & 0.1239 & 0.9814 & 0.9790 & 0.9670 & 0.0809 & 0.9821 & 0.9820 & 0.9794 & 0.6716 & 0.9885 & 0.9857 & 0.9900 \\
      & Embedding $\varepsilon$-Privacy & 0.0368 & 0.9682 & 0.9689 & 0.9576 & 0.0072 & 0.9699 & 0.5827 & 0.9694 & 0.2579 & 0.9973 & 0.9957 & 0.9903 \\
      & Smashed-Data DP                 & 0.0832 & 0.9002 & 0.9283 & 0.9680 & 0.0431 & 0.9777 & 0.9852 & 0.9794 & 0.2805 & 0.9961 & 0.9953 & 0.9894 \\
      & NoPeek                          & 0.0411 & 0.9840 & 0.9843 & 0.9735 & 0.0017 & 0.8612 & 0.1069 & 0.7928 & 0.0974 & 0.9829 & 0.3686 & 0.6169 \\
      & \sysname (ours)                 & \textbf{0.0000} & \textbf{0.0047} & \textbf{0.0036} & \textbf{0.0801} & \textbf{0.0000} & \textbf{0.0000} & \textbf{0.0000} & \textbf{0.0425} & \textbf{0.0006} & \textbf{0.0507} & \textbf{0.0124} & \textbf{0.2658} \\
    \midrule
    \multirow{5}{*}{Ministral-3-8B}
      & No Defense                      & 0.1137 & 0.8672 & 0.8703 & 0.9640 & 0.0508 & 0.9018 & 0.9053 & 0.9810 & 0.7071 & 0.9967 & 0.9971 & 0.9479 \\
      & Embedding $\varepsilon$-Privacy & 0.0639 & 0.8542 & 0.6279 & 0.9216 & 0.0255 & 0.7387 & 0.3046 & 0.9363 & 0.4904 & 0.9938 & 0.9915 & 0.9326 \\
      & Smashed-Data DP                 & 0.0878 & 0.8792 & 0.8925 & 0.9665 & 0.0169 & 0.8474 & 0.8374 & 0.9758 & 0.4712 & 0.9873 & 0.9894 & 0.9491 \\
      & NoPeek                          & 0.1163 & 0.8787 & 0.8798 & 0.9582 & 0.0538 & 0.8916 & 0.4406 & 0.9831 & 0.0640 & 0.2868 & 0.1262 & \textbf{0.1365} \\
      & \sysname (ours)                 & \textbf{0.0016} & \textbf{0.0021} & \textbf{0.0000} & \textbf{0.0000} & \textbf{0.0003} & \textbf{0.0013} & \textbf{0.0000} & \textbf{0.0000} & \textbf{0.0000} & \textbf{0.0016} & \textbf{0.0000} & 0.1807 \\
    \midrule
    \multirow{5}{*}{Ministral-3-14B}
      & No Defense                      & 0.1270 & 0.8843 & 0.8903 & 1.0000 & 0.0786 & 0.9164 & 0.8386 & 0.9993 & 0.6995 & 0.9974 & 0.9977 & 0.9963 \\
      & Embedding $\varepsilon$-Privacy & 0.0316 & 0.8579 & 0.7107 & 0.9494 & 0.0249 & 0.8096 & 0.8000 & 0.9149 & 0.2228 & 0.9953 & 0.9961 & 0.9978 \\
      & Smashed-Data DP                 & 0.0478 & 0.8938 & 0.8996 & 1.0000 & 0.0469 & 0.9435 & 0.5613 & 1.0000 & 0.4142 & 0.9884 & 0.9854 & 0.9913 \\
      & NoPeek                          & 0.0994 & 0.8877 & 0.8903 & 0.8940 & 0.0653 & 0.9225 & 0.7919 & 0.9808 & 0.1939 & 0.9939 & 0.8225 & 0.9941 \\
      & \sysname (ours)                 & \textbf{0.0011} & \textbf{0.0041} & \textbf{0.0000} & \textbf{0.1206} & \textbf{0.0000} & \textbf{0.0025} & \textbf{0.0027} & \textbf{0.1306} & \textbf{0.0014} & \textbf{0.0076} & \textbf{0.0000} & \textbf{0.2730} \\
    \bottomrule
  \end{tabular}%
  }
\end{table*}

\begin{table}[!t]
  \centering
  \caption{Training effectiveness comparison. Banking77 and CLINC150 report classification accuracy. MentalChat16K reports ROUGE-1. Full Model fine-tunes the entire network without a split. Boldface marks the best defense in each column.}
  \label{tab:training}
  \setlength{\tabcolsep}{2.8pt}
  \begin{tabular}{@{}lccc@{}}
    \toprule
    \textbf{Method} & \textbf{Banking77} & \textbf{CLINC150} & \textbf{MentalChat} \\
    \midrule
    \multicolumn{4}{@{}l}{\textbf{Llama-3.2-3B}} \\
    Base & 0.3945 & 0.4453 & 0.3676 \\
    No Defense & 0.6436 & 0.7471 & 0.3916 \\
    Full Model & 0.6445 & 0.7480 & 0.3887 \\
    Emb $\varepsilon$-Priv. & 0.5117 & 0.5557 & 0.3843 \\
    Smashed DP & 0.4883 & 0.5596 & 0.3857 \\
    NoPeek & 0.4482 & 0.6777 & 0.3246 \\
    \sysname (ours) & 0.6221 & 0.7168 & 0.3878 \\
    \sysname (FL) & \textbf{0.6406} & \textbf{0.7236} & \textbf{0.3923} \\
    \midrule
    \multicolumn{4}{@{}l}{\textbf{Llama-3.1-8B}} \\
    Base & 0.5430 & 0.6123 & 0.3626 \\
    No Defense & 0.7090 & 0.8359 & 0.3932 \\
    Full Model & 0.7109 & 0.8340 & 0.4007 \\
    Emb $\varepsilon$-Priv. & 0.6240 & 0.6338 & 0.3837 \\
    Smashed DP & 0.5254 & 0.6016 & \textbf{0.3912} \\
    NoPeek & 0.0000 & 0.0000 & 0.0300 \\
    \sysname (ours) & \textbf{0.6641} & 0.7910 & 0.3911 \\
    \sysname (FL) & 0.6543 & \textbf{0.8018} & 0.3826 \\
    \midrule
    \multicolumn{4}{@{}l}{\textbf{Ministral-3-8B}} \\
    Base & 0.6650 & 0.7109 & 0.2670 \\
    No Defense & 0.7793 & 0.8936 & 0.4038 \\
    Full Model & 0.8037 & 0.8955 & 0.4135 \\
    Emb $\varepsilon$-Priv. & 0.7549 & 0.8623 & 0.3949 \\
    Smashed DP & 0.6699 & 0.7686 & 0.3787 \\
    NoPeek & 0.6699 & 0.8467 & 0.0017 \\
    \sysname (ours) & 0.7588 & 0.8779 & \textbf{0.4062} \\
    \sysname (FL) & \textbf{0.7715} & \textbf{0.8818} & 0.3977 \\
    \midrule
    \multicolumn{4}{@{}l}{\textbf{Ministral-3-14B}} \\
    Base & 0.6758 & 0.7568 & 0.2584 \\
    No Defense & 0.7686 & 0.8672 & 0.3970 \\
    Full Model & 0.7842 & 0.8955 & 0.4080 \\
    Emb $\varepsilon$-Priv. & 0.7246 & 0.8262 & 0.3270 \\
    Smashed DP & 0.7070 & 0.8477 & 0.3755 \\
    NoPeek & 0.7266 & 0.8281 & 0.1210 \\
    \sysname (ours) & \textbf{0.7510} & 0.8760 & \textbf{0.4003} \\
    \sysname (FL) & \textbf{0.7510} & \textbf{0.8789} & 0.3800 \\
    \bottomrule
  \end{tabular}
\end{table}

\subsection{Experimental Setup}

\noindent\textbf{Models.}
We evaluate four instruction-tuned LLMs. The model suite includes Llama-3.2-3B-Instruct and Llama-3.1-8B-Instruct~\cite{grattafiori2024llama3herdmodels}, together with Ministral-3-8B-Instruct and Ministral-3-14B-Instruct~\cite{liu2026ministral3}. These models span two model families and parameter counts from 3B to 14B. We evaluate downstream task performance for every model and dataset combination.

\noindent\textbf{Datasets.}
We evaluate three downstream datasets. Banking77~\cite{Casanueva2020_banking77} contains online banking requests labeled with 77 fine-grained customer intents. CLINC150~\cite{larson-etal-2019-evaluation} covers 150 intent classes across multiple service domains. MentalChat16K~\cite{MentalChat16K} contains mental health dialogues for counseling-style response generation. Together, these three datasets span single-domain classification, multi-domain classification, and open-ended dialogue generation, letting us test the adapter protocol across different task types and domains. The client's input text is sensitive in all three cases, whether it is a banking request, a service inquiry, or a personal dialogue, and we evaluate both defense capability and training effectiveness on all three datasets. All training, evaluation, and alignment subsets used within each downstream task are non-overlapping. In the FL experiments, the private training data are partitioned into non-overlapping local subsets across clients. For the alignment dataset $\mathcal{D}_{\mathrm{align}}$ used to train $W_1$ and $W_2$ (Section~\ref{sec:adapter_training}), we use OASST1~\cite{oasst1}, a crowd-sourced corpus of multi-turn assistant conversations. OASST1 is unrelated to Banking77, CLINC150, and MentalChat16K in both task and domain, which lets us verify that the adapters can be trained without any access to data resembling the private fine-tuning task.

\noindent\textbf{Baselines.}
We compare \sysname against three baseline defenses, all reproduced under the same SL pipeline.

\textit{Embedding $\varepsilon$-Privacy}~\cite{DP_attack} applies noise at the embedding layer. A unit norm direction $\mathbf{v} = \mathcal{N}(0,I_d)/\|\mathcal{N}(0,I_d)\|$ and a magnitude $r \sim \mathrm{Gamma}(d,\, \varepsilon d)$ are sampled, where the second parameter is the rate. The noisy embedding $\tilde{e} = e + r\mathbf{v}$ is then mapped to the vocabulary token with the largest embedding dot product, replacing the noisy embedding with the corresponding token embedding.

\textit{Smashed-Data DP}~\cite{DP_forward_attack} applies Gaussian noise directly to the smashed data, as follows.
\begin{equation}
s'_A = s_A + \sigma \cdot \mathcal{N}(0, I)
\end{equation}
where $\sigma$ controls the noise magnitude.

\textit{NoPeek}~\cite{NoPeak} adds a distance correlation regularization term to the training objective, as follows.
\begin{equation}
\mathcal{L} = \mathcal{L}_{\mathrm{CE}} + \lambda \cdot \mathrm{dCor}(e,\, s_A)
\end{equation}
where $e$ denotes the input embedding, $s_A$ the smashed data, and $\lambda$ the regularization weight.

Table~\ref{tab:baseline_params} in Appendix~\ref{sec:appendix_hyperparams} lists the hyperparameter values used for each baseline. Each of these hyperparameters governs the same privacy-utility trade-off as $\lambda_{W_2}^{(1)}$ does in our method, so we tuned them carefully. Appendix~\ref{sec:appendix_baseline_tuning} describes the tuning procedure and reports results across the tuning range. We additionally include \textit{No Defense} as a privacy and utility reference. All split methods in this evaluation, including No Defense, the three baselines, and \sysname, freeze $B$ and update only $A$ and $C$. \textit{Full Model} fine-tunes the entire network without a split and is reported as a utility reference. 

\noindent\textbf{Defense capability setup.}
To evaluate defense against DRA, we follow prior work~\cite{CCS_unveiling} and adopt the worst-case single-client setting, in which the server has full access to all smashed data and the model parameters from a client with no interference from other participants. This access excludes the client-side adapters $W_1$ and $W_2$, which remain local under our threat model. The setup maximizes the adversary's advantage within our stated scope and provides a strong attack surface for evaluating \sysname's privacy protection.

\noindent\textbf{Training effectiveness setup.}
In addition to evaluating training effectiveness in the single-client setting, we also measure it under FL conditions. For this, we use 64 clients, each holding its own local dataset partition, randomly assigned into 8 groups that each share an adapter pair $(W_1, W_2)$, emulating institutions in which multiple client devices operate under a common adapter policy. Each adapter pair remains local to its group throughout training and is never shared with the server or with clients outside the group. In each of 256 global rounds, 8 clients are selected to each perform 4 local steps, after which the server aggregates the shared $A$ and $C$ parameters across the selected clients via FedAvg. The alignment data used by each group is also non-overlapping with that of every other group.

\begin{figure}[t]
    \centering
    \includegraphics[width=0.8\linewidth]{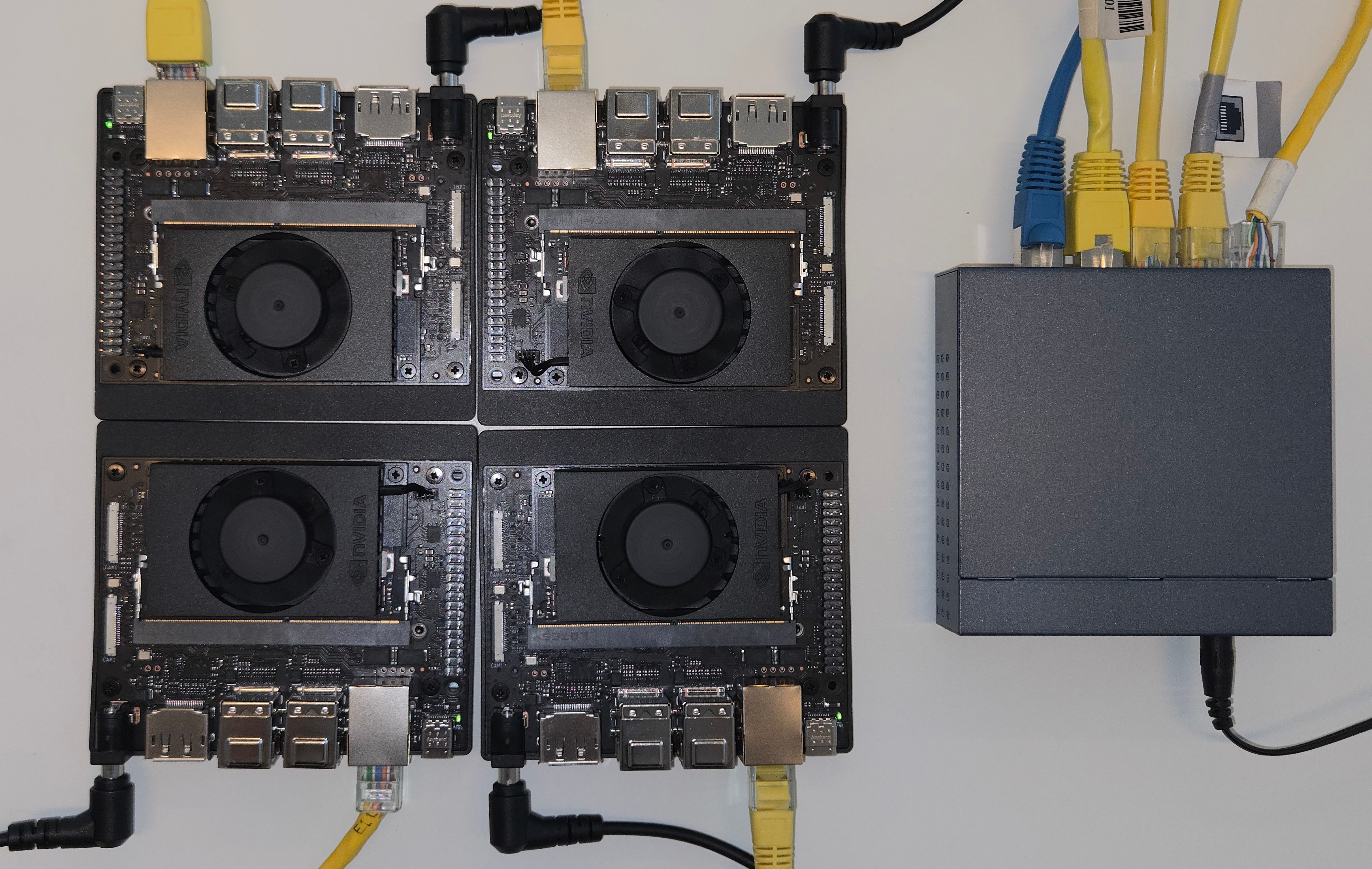}
    \caption{Physical testbed used to evaluate \sysname. With \sysname, edge devices such as the NVIDIA Jetson Orin Nano can serve as clients in FL-based LLM fine-tuning.}
    \label{fig:testbed}
\end{figure}

\noindent\textbf{Hyperparameters and hardware.}
Table~\ref{tab:hyperparams} in Appendix~\ref{sec:appendix_hyperparams} summarizes the hyperparameters used across all experiments, including the learning rate settings that vary by dataset and the per-model adapter coupling weight. The main experiments are conducted on an NVIDIA RTX PRO 6000 with 96~GB GPU memory. Client-side overhead, including adapter inference and memory consumption during fine-tuning, is measured on an NVIDIA Jetson Orin Nano with 8~GB unified memory to reflect edge device constraints, as shown in Figure~\ref{fig:testbed}.

\subsection{Evaluation Methodology}

We evaluate \sysname from three perspectives.

\noindent\textbf{Defense capability.}
We assess the privacy protection provided by \sysname by measuring how well an adversary can reconstruct the client's private training data. We evaluate against four attacks, \textit{Direct Forward}, \textit{TAG}, \textit{LAMP}, and \textit{BiSR}. Under Direct Forward, the server directly applies $f_C \circ f_B$ to the received smashed data to decode the output. This is the most natural and zero-cost attack available to the server. TAG, LAMP, and BiSR correspond to stronger adversaries that rely on optimization or search and were originally designed for gradient- and embedding-based reconstruction attacks against LLMs. BiSR's inversion decoder is trained on $A$'s output $s_A$ on an auxiliary dataset. Under our defense, we do not retrain this decoder on $W_1$'s output $s'_A$. Doing so would require the adversary to produce paired examples of known inputs and the corresponding $s'_A$ on an auxiliary dataset, but $W_1$ never leaves the client. For a fair comparison, we also apply the same $A$-trained inversion decoder when evaluating BiSR against the other baselines. We report the mean ROUGE-1 of the reconstructed private inputs as the primary reconstruction quality metric.

All four attacks in Table~\ref{tab:defense} are measured at the final checkpoint. Direct Forward is also tracked throughout training. TAG, LAMP, and BiSR are computationally intensive and are evaluated at the final checkpoint only.

\noindent\textbf{Training effectiveness.}
We measure whether the adapter protocol preserves the utility of fine-tuning. Since Banking77 and CLINC150 are intent classification tasks, we report classification accuracy on their evaluation sets. Since MentalChat16K is a free-form generation task, we report ROUGE-1 on its evaluation set. We compare the pre-trained base model, plain SL without privacy protection, full-model fine-tuning, the baselines, \sysname in the single-client setting, and \sysname in the FL setting under these metrics.

\noindent\textbf{Overhead.}
We measure two overhead metrics. The first is the per-step client-side computation time and memory footprint introduced by running $W_1$ and $W_2$ during fine-tuning, relative to the plain SL baseline. The second is the end-to-end wall-clock training time.

\subsection{Defense Capability}
\label{sec:exp_defense}

Table~\ref{tab:defense} reports the ROUGE-1 of reconstructed private inputs under each attack. Without any defense, Direct Forward attains high reconstruction quality on MentalChat16K, showing that the server's available decoder poses an immediate threat on long-form inputs. Direct Forward scores are lower on Banking77 and CLINC150 because the private texts are short, but TAG, LAMP, and BiSR remain high on those datasets as well, so standard SL still exhibits substantial reconstruction leakage under stronger attacks.

The baselines can reduce Direct Forward scores, but TAG, LAMP, and BiSR remain high in most settings. Some NoPeek entries show lower attack scores together with collapsed training utility in Table~\ref{tab:training}.

\sysname substantially reduces reconstruction scores. Direct Forward, TAG, and LAMP stay near zero. BiSR remains the most difficult attack for \sysname to suppress. Across the evaluated settings, \sysname substantially outperforms the baselines in the vast majority of attack columns. 

\subsection{Training Effectiveness}
\label{sec:exp_training}

Table~\ref{tab:training} reports classification accuracy on Banking77 and CLINC150, and ROUGE-1 on MentalChat16K. Among the baselines, training effectiveness varies across methods, models, and datasets. Embedding $\varepsilon$-Privacy and Smashed-Data DP often fall below the No Defense split baseline, especially on the classification tasks. NoPeek collapses on several MentalChat16K settings.

\sysname achieves training effectiveness close to the no-defense references across models and datasets. The results for \sysname (FL) are comparable to the single-client setting. This indicates that the per-client adapter design can scale to federated conditions in our setup without clear degradation from client-specific adapters.

Taken together with Table~\ref{tab:defense}, on the vast majority of model-dataset combinations \sysname simultaneously provides stronger privacy protection than every baseline and matches or exceeds their training effectiveness, rather than trading one for the other. In contrast, the baselines in Table~\ref{tab:training} that come closest to \sysname's privacy protection, such as NoPeek on several MentalChat16K settings, do so only by collapsing training utility.

\subsection{Direct Forward Attack During Training}
\label{sec:exp_direct_fwd_curve}

\begin{figure*}[t]
    \centering
    \subfloat[Llama-3.2-3B]{\includegraphics[width=0.24\textwidth]{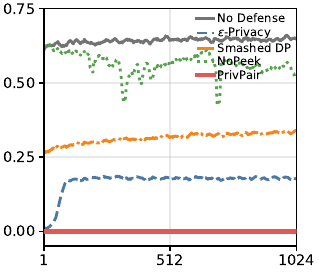}}
    \hfill
    \subfloat[Llama-3.1-8B]{\includegraphics[width=0.24\textwidth]{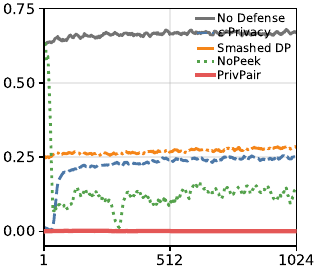}}
    \hfill
    \subfloat[Ministral-3-8B]{\includegraphics[width=0.24\textwidth]{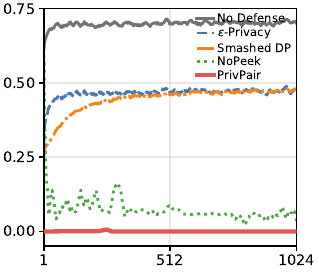}}
    \hfill
    \subfloat[Ministral-3-14B]{\includegraphics[width=0.24\textwidth]{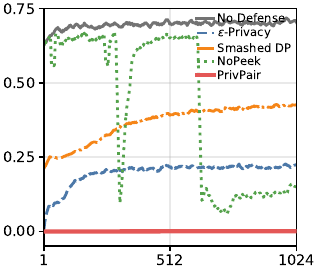}}
    \caption{Sixteen-step moving mean Direct Forward ROUGE-1 on MentalChat16K throughout training under different defense conditions.}
    \label{fig:rl_curve}
\end{figure*}

Figure~\ref{fig:rl_curve} shows how the Direct Forward ROUGE-1 evolves over the course of fine-tuning on MentalChat16K for every model in the model suite. For No Defense, Embedding $\varepsilon$-Privacy, and Smashed-Data DP, the reconstruction quality rises as training progresses. This increase follows from the fine-tuning objective, since training the LLM to predict the next token given the smashed data also improves the decoder available to the server for the client's input. Defenses that rely solely on perturbing the smashed data are limited by this effect, because the perturbation must remain bounded to preserve training utility while the server's available decoder adapts over time. \sysname stays near zero on every model throughout training, showing that the obfuscate-and-recover scheme remains stable as fine-tuning proceeds.

\subsection{Overhead}
\label{sec:exp_overhead}

\subsubsection{Training Overhead}
\label{sec:exp_training_overhead}

The adapter pair $W_1$ and $W_2$ introduces only a small additional memory footprint relative to the base client-side model. On the NVIDIA Jetson Orin Nano, the $A$ and $C$ components of Llama-3.2-3B occupy 2{,}655~MB of GPU memory, while the two adapters together require only 144~MB, amounting to approximately 5.4\% of the base memory cost. Furthermore, since the adapters do not participate in the main computation graph under the STE protocol, they do not incur additional activation memory during backpropagation.
We measure the per-step client-side computation time on the NVIDIA Jetson Orin Nano, evaluating only the client-side components $A$ and $C$. Processing 16 samples (sequence length 64, batch size 1) takes 3.70~seconds without adapters and 3.74~seconds with adapters. The small latency increase indicates that running $W_1$ and $W_2$ on a resource-constrained edge device introduces limited computational overhead during fine-tuning. During the main fine-tuning steps, the smashed data transmitted to the server remains a single tensor of the same shape as in the plain SL baseline, and the model update upload cost is unchanged, so running $W_1$ and $W_2$ on the client introduces no additional communication overhead beyond the alignment cost.

\subsubsection{End-to-End Training Time}
\label{sec:exp_e2e_time}

\begin{table*}[!ht]
  \centering
  \caption{Measured wall-clock training time, all on a single GPU. Percentages report the overhead relative to No Defense. \sysname (train) is the main fine-tuning loop, including recurring burst alignment. \sysname (init align) is the one-time initial alignment cost. \sysname (total) sums the two.}
  \label{tab:training_time}
  \small
  \resizebox{\textwidth}{!}{%
  \begin{tabular}{lccccccc}
    \toprule
    \textbf{Model} & \textbf{No Defense} & \textbf{Emb $\varepsilon$-Priv.} & \textbf{Smashed DP} & \textbf{NoPeek} & \textbf{\sysname (train)} & \textbf{\sysname (init align)} & \textbf{\sysname (total)} \\
    \midrule
    Llama-3.2-3B    & 13.84 min & 16.90 min (+22.1\%) & 14.10 min (+1.9\%)  & 15.57 min (+12.5\%) & 14.50 min (+4.8\%)  & 4.80 min  & 19.30 min (+39.5\%) \\
    Llama-3.1-8B    & 27.61 min & 31.43 min (+13.8\%) & 27.18 min (-1.5\%)  & 29.55 min (+7.0\%)  & 29.20 min (+5.8\%)  & 8.95 min  & 38.16 min (+38.2\%) \\
    Ministral-3-8B  & 29.12 min & 33.08 min (+13.6\%) & 28.70 min (-1.4\%)  & 31.00 min (+6.5\%)  & 30.72 min (+5.5\%)  & 9.42 min  & 40.14 min (+37.8\%) \\
    Ministral-3-14B & 50.26 min & 61.67 min (+22.7\%) & 51.00 min (+1.5\%)  & 57.07 min (+13.6\%) & 53.78 min (+7.0\%)  & 16.05 min & 69.83 min (+38.9\%) \\
    \bottomrule
  \end{tabular}%
  }
\end{table*}

Table~\ref{tab:training_time} reports the measured wall-clock training time on MentalChat16K. Excluding the one-time initial alignment cost, \sysname's main training loop, which already includes the recurring burst alignment steps, is 5\% to 10\% slower than No Defense, a range comparable to or smaller than the overhead of NoPeek and Embedding $\varepsilon$-Privacy. Smashed-Data DP adds only a fixed noise term to the smashed data and, as expected, shows close to zero overhead. Including the one-time initial alignment cost, \sysname's total training time is around 40\% higher than No Defense, and this share shrinks as the fine-tuning loop grows longer, since the initial cost does not scale with the number of training steps.

\section{Conclusion}
\label{sec:conclusion}

We presented \sysname, a privacy-preserving framework for LLM fine-tuning based on federated split learning that addresses the tension between smashed-data privacy and trainability. We showed that the autoregressive nature of LLMs makes perturbation-based defenses insufficient in this setting, since fine-tuning can improve the decoder available to the server while the perturbation must remain bounded to preserve utility. Our core contribution is a learned obfuscate-and-recover protocol implemented by two lightweight client-side adapters. $W_1$ makes the transmitted activations less compatible with the server's decoder, while $W_2$ maps the returned activations toward a representation compatible with the client-side model. The adapters remain local to the client because they are never uploaded. An STE-based backward pass preserves effective updates to the deployable model. Our analysis characterizes the trainability conditions of the adapter protocol and shows how training quality depends on the achieved obfuscation and recovery errors. Our defense evaluation across four LLMs and three downstream datasets demonstrates that \sysname substantially reduces reconstruction quality under the evaluated attacks while maintaining utility close to the no-defense references.

%\section*{Acknowledgment}

\bibliographystyle{IEEEtran}
\bibliography{IEEEabrv,ref}

\appendices
\section{Hyperparameters}
\label{sec:appendix_hyperparams}

Table~\ref{tab:hyperparams} summarizes the hyperparameters used across all experiments. A few settings vary by model or dataset and are described here instead of in the table.

We set $A$ to contain 5 transformer layers. Modern LLMs adopt a residual transformer architecture, and this gives the smashed data several layers of transformation beyond the raw input embeddings, so trivial reconstruction by attaching the language model head directly to $s_A$ is not possible under this split. We set $C$ to contain 1 layer.

We also vary the learning rate by dataset. Banking77 and CLINC150 are classification tasks in which the assistant response is a short label, so we set the learning rate to $1\times10^{-5}$ for these two datasets. MentalChat16K requires generating a long, free-form assistant response, so we set the learning rate to $5\times10^{-5}$ for this dataset.

When computing training effectiveness, we also vary the evaluation set size and the generation length by dataset. For Banking77 and CLINC150, we evaluate on 1{,}024 samples and generate at most 64 tokens per sample, matching the short label outputs of these classification tasks. For MentalChat16K, we evaluate on 256 samples and generate at most 256 tokens per sample, matching the longer free-form assistant responses of this dataset.

\begin{table}[H]
\centering
\caption{Hyperparameter settings used in all experiments.}
\label{tab:hyperparams}
\begin{tabular}{lll}
\toprule
\textbf{Category} & \textbf{Parameter} & \textbf{Value} \\
\midrule
\multirow{2}{*}{$W_1$ adapter}
  & Layer num.               & 2 \\
  & Hidden dimension         & $1\times d_{\mathrm{model}}$ \\
\midrule
\multirow{2}{*}{$W_2$ adapter}
  & Layer num.               & 2 \\
  & Hidden dimension         & $2\times d_{\mathrm{model}}$ \\
\midrule
\multirow{2}{*}{$W_1$ loss}
  & $\lambda_{\mathrm{CKA}}^{(1)}$  & 1.0 \\
  & $\lambda_{\mathrm{JSD}}^{(1)}$  & 2.0 \\
\midrule
\multirow{2}{*}{$W_2$ loss}
  & $\lambda_{\mathrm{KL}}^{(2)}$   & 1.0 \\
  & $\lambda_{\mathrm{MSE}}^{(2)}$  & 2.0 \\
\midrule
\multirow{10}{*}{Training}
  & $W_1$ adapter learning rate  & $2\times10^{-4}$ \\
  & $W_2$ adapter learning rate  & $4\times10^{-4}$ \\
  & Adapter LR warmup steps      & 32 \\
  & Adapter LR eta min ratio     & 0.1 \\
  & Batch size           & 8 \\
  & Max sequence length  & 768 \\
  & Training steps       & 1{,}024 \\
  & Alignment steps & 256 \\
  & Burst interval & 256 training steps \\
  & Burst alignment steps & 16 \\
\bottomrule
\end{tabular}
\end{table}

$\lambda_{W_2}^{(1)}$ is also a coefficient in the $W_1$ loss, but its value is tuned separately for each model, as it depends on the model's response to $W_1$ obfuscation during alignment. Table~\ref{tab:aw2w_per_model} lists the value used for each model.

\begin{table}[H]
\centering
\caption{Per-model value of $\lambda_{W_2}^{(1)}$.}
\label{tab:aw2w_per_model}
\begin{tabular}{lc}
\toprule
\textbf{Model} & $\lambda_{W_2}^{(1)}$ \\
\midrule
Llama-3.2-3B    & 1.0 \\
Llama-3.1-8B    & 0.6 \\
Ministral-3-8B  & 1.4 \\
Ministral-3-14B & 1.4 \\
\bottomrule
\end{tabular}
\end{table}

Table~\ref{tab:baseline_params} lists the operating points used for the baselines on each model and dataset. $\varepsilon$ is the Embedding $\varepsilon$-Privacy scale, $\sigma$ is the Smashed-Data DP scale, and $\lambda$ is the NoPeek scale.

\begin{table}[H]
\centering
\caption{Selected operating points used for the baselines in the main results.}
\label{tab:baseline_params}
\footnotesize
\setlength{\tabcolsep}{3.5pt}
\begin{tabular}{llccc}
\toprule
\textbf{Model} & \textbf{Dataset} & $\varepsilon$ & $\sigma$ & $\lambda$ \\
\midrule
\multirow{3}{*}{Llama-3.2-3B}
  & Banking & 0.1  & 0.2  & $1\times10^{2}$ \\
  & CLINC   & 0.1  & 0.2  & $5\times10^{1}$ \\
  & Mental  & 0.05 & 0.6  & $1\times10^{2}$ \\
\midrule
\multirow{3}{*}{Llama-3.1-8B}
  & Banking & 0.2  & 0.2  & $5\times10^{1}$ \\
  & CLINC   & 0.2  & 0.2  & $1\times10^{2}$ \\
  & Mental  & 0.1  & 0.54 & $1\times10^{2}$ \\
\midrule
\multirow{3}{*}{Ministral-3-8B}
  & Banking & 0.3  & 0.2  & $2.5\times10^{1}$ \\
  & CLINC   & 0.3  & 0.2  & $1\times10^{1}$ \\
  & Mental  & 0.2  & 0.6  & $1\times10^{2}$ \\
\midrule
\multirow{3}{*}{Ministral-3-14B}
  & Banking & 0.2  & 0.2  & $1\times10^{1}$ \\
  & CLINC   & 0.2  & 0.2  & $1\times10^{1}$ \\
  & Mental  & 0.1  & 1    & $1\times10^{1}$ \\
\bottomrule
\end{tabular}
\end{table}

We implement TAG, LAMP, and BiSR following the official implementation of BiSR~\cite{CCS_unveiling}. Direct Forward has no additional hyperparameters. All remaining attack hyperparameters follow the default settings of that implementation. Table~\ref{tab:attack_hyperparams} lists these values.

\begin{table}[H]
\centering
\caption{Default attack hyperparameters used in all experiments.}
\label{tab:attack_hyperparams}
\begin{tabular}{lll}
\toprule
\textbf{Category} & \textbf{Parameter} & \textbf{Value} \\
\midrule
\multirow{4}{*}{TAG}
  & Epochs               & 300 \\
  & $\beta$              & 0.85 \\
  & Learning rate        & $0.09$ \\
  & Initialization temperature & $1.0$ \\
\midrule
\multirow{4}{*}{LAMP}
  & Epochs               & 300 \\
  & $\beta$              & 0.85 \\
  & Learning rate        & $0.09$ \\
  & Reorder frequency    & 30 \\
\midrule
\multirow{3}{*}{BiSR matching}
  & Epochs               & 20 \\
  & Learning rate        & $1\times10^{-4}$ \\
  & Weight decay         & $0.01$ \\
\midrule
\multirow{7}{*}{BiSR inverter}
  & Hidden size          & 256 \\
  & Dropout              & $0.1$ \\
  & Learning rate        & $1\times10^{-3}$ \\
  & Weight decay         & $1\times10^{-5}$ \\
  & Epochs               & 20 \\
  & Batch size           & 6 \\
\bottomrule
\end{tabular}
\end{table}

The BiSR inverter is trained separately for each model and each downstream task. We reserve 512 samples from each dataset as an extra auxiliary dataset for this training. This auxiliary dataset does not overlap with any other split used in the experiments.

\section{Baseline Hyperparameter Tuning}
\label{sec:appendix_baseline_tuning}

Each baseline exposes a hyperparameter that governs a privacy-utility trade-off, so a single fixed setting of this hyperparameter can make the baseline look arbitrarily weak along one axis while looking strong along the other. Comparing \sysname against a baseline evaluated at only one such setting is therefore not sufficient to establish that \sysname outperforms the baseline. To make this comparison meaningful, for each baseline we sweep its hyperparameter over a wide range and show that at some point in this range, the baseline is simultaneously no better than \sysname on both privacy protection and training effectiveness. This rules out the possibility that the baseline's apparent weakness in our main results is an artifact of an unfavorable hyperparameter choice.

\begin{figure}[t]
    \centering
    \subfloat[Smashed-Data DP]{\includegraphics[width=0.95\linewidth]{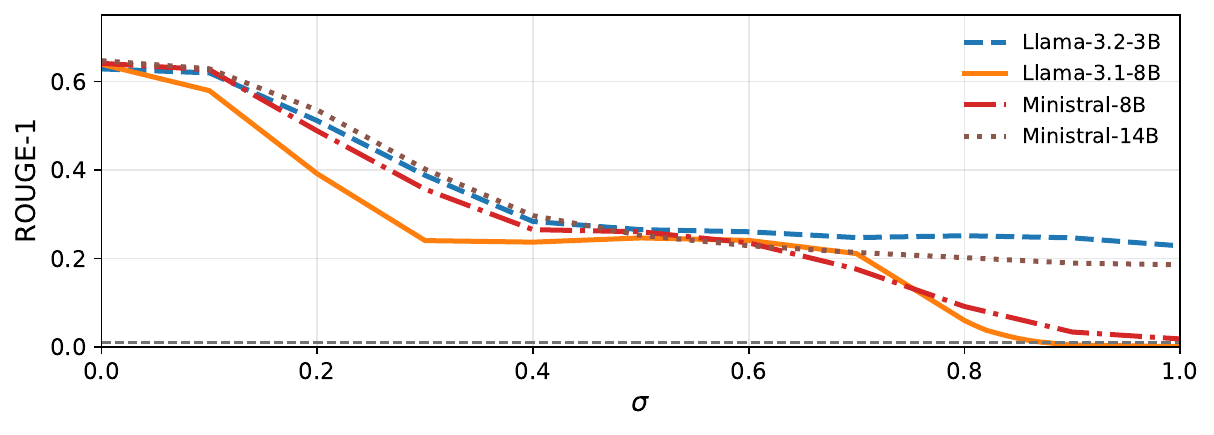}\label{fig:scan_dp}}\\
    \subfloat[Embedding $\varepsilon$-Privacy]{\includegraphics[width=0.95\linewidth]{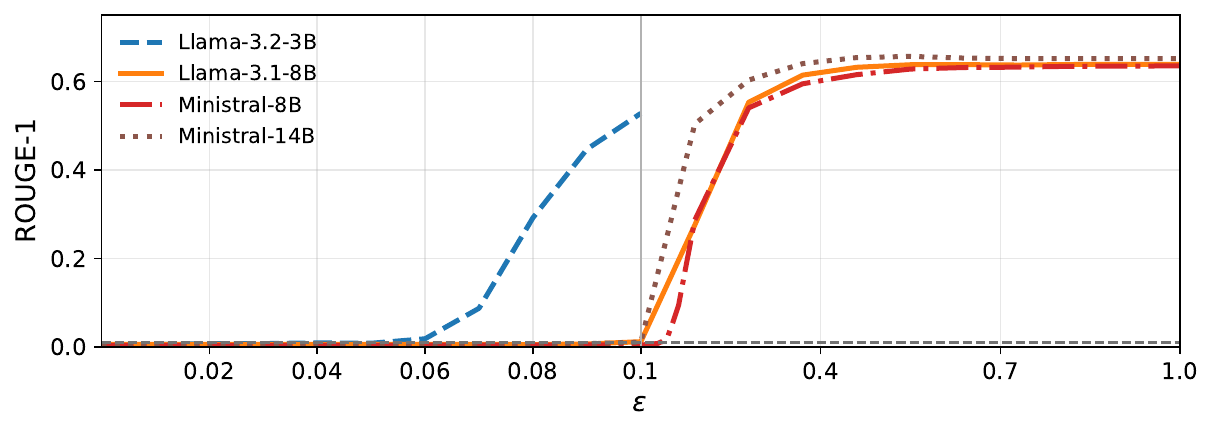}\label{fig:scan_dxp}}\\
    \subfloat[NoPeek]{\includegraphics[width=0.95\linewidth]{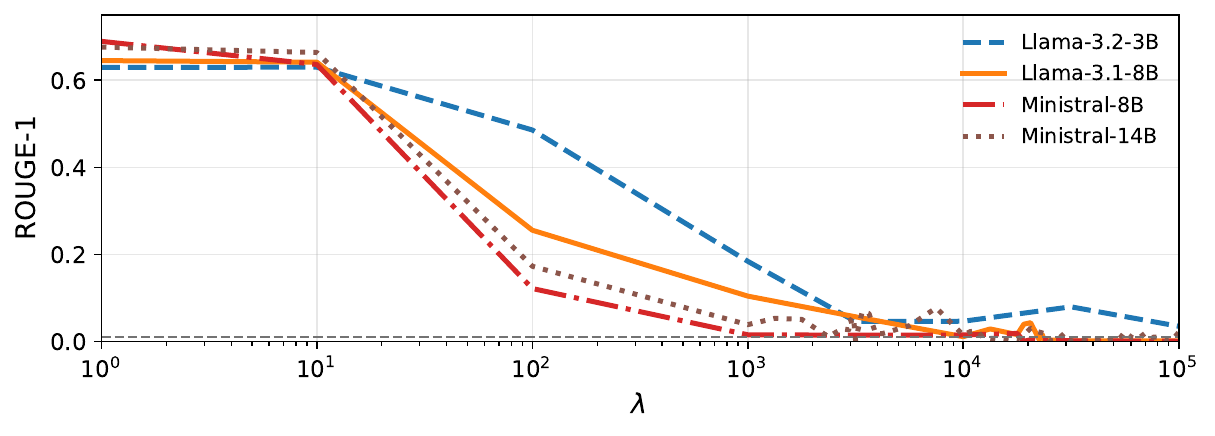}\label{fig:scan_nopeek}}
    \caption{Direct Forward attack ROUGE-1 on MentalChat16K across the scanned hyperparameter range for each baseline, shown for every model in the model suite. The onset, midpoint, and reach operating points described in this section are selected from these curves.}
    \label{fig:rouge1_scan}
\end{figure}

Figure~\ref{fig:rouge1_scan} shows the resulting scan for each baseline. For Smashed-Data DP and Embedding $\varepsilon$-Privacy, we sweep a wide range of hyperparameter values directly on the base model and record the Direct Forward attack ROUGE-1 at each value. From this sweep we select three operating points. The first point is where the attack ROUGE-1 first drops noticeably, indicating that the defense has begun to take effect, and we treat this point as a lower bound on privacy. The second point is where the attack ROUGE-1 reaches 0.01, and we treat this point as an upper bound on privacy. Utility is not known at this stage, so these points are selected from the attack curve alone. The third point is the midpoint between the first and second points. For Embedding $\varepsilon$-Privacy and Smashed-Data DP this midpoint is the arithmetic mean. For NoPeek the midpoint is the rounded geometric mean. Because the NoPeek defense is applied as part of training and only becomes effective after training has proceeded for some time, we record the attack ROUGE-1 over the full training run and select three points from this curve using the same procedure. We carry out this tuning procedure on MentalChat16K and tune the hyperparameter separately for each model.

Tables~\ref{tab:key_scales_dxp}, \ref{tab:key_scales_dp}, and \ref{tab:key_scales_nopeek} list the resulting onset, midpoint, and reach scales for each model. For some models the Direct Forward ROUGE-1 never falls to 0.01 within the scanned range, and in those cases reach is the strongest scale in the scan.

\begin{table}[H]
\centering
\caption{Onset, midpoint, and reach values of $\varepsilon$ for Embedding $\varepsilon$-Privacy on MentalChat16K.}
\label{tab:key_scales_dxp}
\begin{tabular}{lccc}
\toprule
\textbf{Model} & \textbf{Onset} & \textbf{Midpoint} & \textbf{Reach} \\
\midrule
Llama-3.2-3B    & 0.1 & 0.075 & 0.05 \\
Llama-3.1-8B    & 0.2 & 0.15  & 0.1  \\
Ministral-3-8B  & 0.3 & 0.2   & 0.1  \\
Ministral-3-14B & 0.2 & 0.15  & 0.1  \\
\bottomrule
\end{tabular}
\end{table}

\begin{table}[H]
\centering
\caption{Onset, midpoint, and reach values of $\sigma$ for Smashed-Data DP on MentalChat16K.}
\label{tab:key_scales_dp}
\begin{tabular}{lccc}
\toprule
\textbf{Model} & \textbf{Onset} & \textbf{Midpoint} & \textbf{Reach} \\
\midrule
Llama-3.2-3B    & 0.2 & 0.6  & 1    \\
Llama-3.1-8B    & 0.2 & 0.54 & 0.87 \\
Ministral-3-8B  & 0.2 & 0.6  & 1    \\
Ministral-3-14B & 0.2 & 0.6  & 1    \\
\bottomrule
\end{tabular}
\end{table}

\begin{table}[H]
\centering
\caption{Onset, midpoint, and reach values of $\lambda$ for NoPeek on MentalChat16K.}
\label{tab:key_scales_nopeek}
\begin{tabular}{lccc}
\toprule
\textbf{Model} & \textbf{Onset} & \textbf{Midpoint} & \textbf{Reach} \\
\midrule
Llama-3.2-3B    & $1\times10^{2}$ & $3\times10^{3}$ & $1\times10^{5}$ \\
Llama-3.1-8B    & $1\times10^{2}$ & $1\times10^{3}$ & $1\times10^{4}$ \\
Ministral-3-8B  & $1\times10^{2}$ & $1\times10^{3}$ & $2\times10^{4}$ \\
Ministral-3-14B & $1\times10^{1}$ & $2\times10^{2}$ & $3\times10^{3}$ \\
\bottomrule
\end{tabular}
\end{table}

For NoPeek, the three operating points follow the same onset, midpoint, and reach rules as the other baselines. In practice, the midpoint and reach values of $\lambda$ are often so large that training collapses completely. When all three scanned points fail in this way, we decrease $\lambda$ and use a better scale as the selected operating point for that model and dataset.

\begin{figure*}[t]
    \centering
    \subfloat[Banking77]{\includegraphics[width=0.24\textwidth]{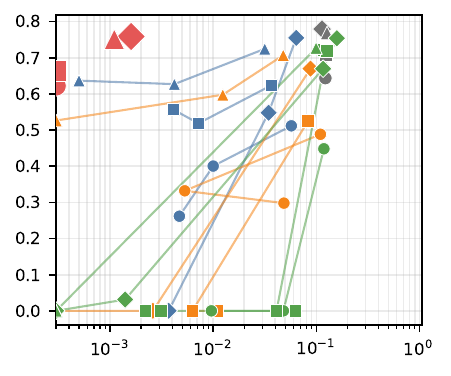}}
    \hfill
    \subfloat[CLINC150]{\includegraphics[width=0.24\textwidth]{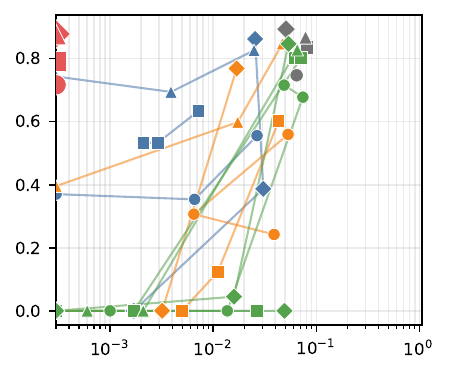}}
    \hfill
    \subfloat[MentalChat16K]{\includegraphics[width=0.24\textwidth]{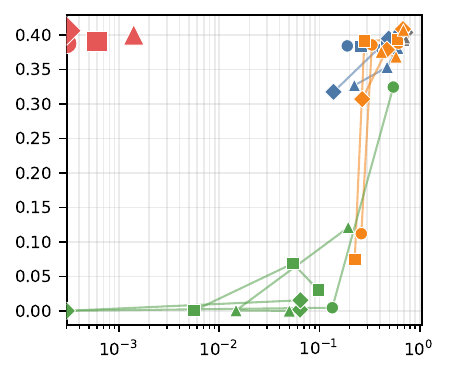}}
    \hfill
    \subfloat{\includegraphics[width=0.18\textwidth]{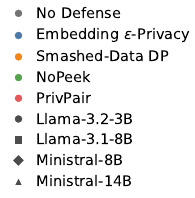}}
    \caption{Privacy-utility landscape of the scanned baseline scales, No Defense, and \sysname. The horizontal axis is Direct Forward ROUGE-1 at the final checkpoint, shown on a log scale. The vertical axis is task utility. The top left is better, with lower leakage and higher utility.}
    \label{fig:pareto}
\end{figure*}

Figure~\ref{fig:pareto} places the scanned baseline scales, No Defense, and \sysname in the privacy-utility plane. Utility is classification accuracy on Banking77 and CLINC150 and task ROUGE-1 on MentalChat16K. Direct Forward ROUGE-1 is measured at the final checkpoint. Color denotes the defense method, and marker shape denotes the model.

Tables~\ref{tab:unselected_dxp}, \ref{tab:unselected_dp}, and \ref{tab:unselected_nopeek} report training effectiveness and Direct Forward leakage for every scanned operating point that is not the selected point used in the main results. Gray cells mark collapsed training, including near-zero task scores below $0.1$. Light red cells mark points that are worse than \sysname on both axes, with a lower task score and a higher Direct Forward score. A yellow task-score cell marks a non-collapsed point whose accuracy or task ROUGE-1 is below the corresponding base model. A light green task-score cell marks a point that is better than \sysname on this axis.

\newcommand{\gcells}[3]{\cellcolor{gray!25}#1 & \cellcolor{gray!25}#2 & \cellcolor{gray!25}#3}
\newcommand{\rcells}[3]{\cellcolor{red!15}#1 & \cellcolor{red!15}#2 & \cellcolor{red!15}#3}
\newcommand{\rycells}[3]{\cellcolor{red!15}#1 & \cellcolor{yellow!40}#2 & \cellcolor{red!15}#3}
\newcommand{\ugcells}[3]{#1 & \cellcolor{green!20}#2 & #3}

\begin{table}[ht]
  \centering
  \caption{Unselected Embedding $\varepsilon$-Privacy operating points.}
  \label{tab:unselected_dxp}
  \footnotesize
  \setlength{\tabcolsep}{3.5pt}
  \begin{tabular}{llrrr}
    \toprule
    \textbf{Model} & \textbf{Dataset} & $\varepsilon$ & \makecell{\textbf{Acc. /}\\\textbf{ROUGE-1}} & \makecell{\textbf{Direct Fwd}\\\textbf{ROUGE-1}} \\
    \midrule
    \multirow{6}{*}{Llama-3.2-3B}
      & \multirow{2}{*}{Banking} & \rcells{0.075}{0.4004}{0.0100} \\
      & & \rycells{0.05}{0.2617}{0.0047} \\
      \cmidrule{2-5}
      & \multirow{2}{*}{CLINC} & \rycells{0.075}{0.3535}{0.0066} \\
      & & \rycells{0.05}{0.3701}{0.0002} \\
      \cmidrule{2-5}
      & \multirow{2}{*}{Mental} & \ugcells{0.1}{0.3928}{0.5721} \\
      & & \rcells{0.075}{0.3797}{0.4206} \\
    \midrule
    \multirow{6}{*}{Llama-3.1-8B}
      & \multirow{2}{*}{Banking} & \rycells{0.15}{0.5186}{0.0072} \\
      & & \rcells{0.1}{0.5566}{0.0041} \\
      \cmidrule{2-5}
      & \multirow{2}{*}{CLINC} & \rycells{0.15}{0.5332}{0.0029} \\
      & & \rycells{0.1}{0.5332}{0.0021} \\
      \cmidrule{2-5}
      & \multirow{2}{*}{Mental} & \rcells{0.2}{0.3841}{0.5504} \\
      & & \rcells{0.15}{0.3832}{0.4655} \\
    \midrule
    \multirow{6}{*}{Ministral-3-8B}
      & \multirow{2}{*}{Banking} & \rycells{0.2}{0.5479}{0.0344} \\
      & & \gcells{0.1}{0.0000}{0.0037} \\
      \cmidrule{2-5}
      & \multirow{2}{*}{CLINC} & \rycells{0.2}{0.3867}{0.0306} \\
      & & \gcells{0.1}{0.0000}{0.0017} \\
      \cmidrule{2-5}
      & \multirow{2}{*}{Mental} & \rcells{0.3}{0.4041}{0.6501} \\
      & & \rcells{0.1}{0.3175}{0.1380} \\
    \midrule
    \multirow{6}{*}{Ministral-3-14B}
      & \multirow{2}{*}{Banking} & \rycells{0.15}{0.6270}{0.0042} \\
      & & \rycells{0.1}{0.6367}{0.0005} \\
      \cmidrule{2-5}
      & \multirow{2}{*}{CLINC} & \rycells{0.15}{0.6943}{0.0039} \\
      & & \rycells{0.1}{0.7432}{0.0000} \\
      \cmidrule{2-5}
      & \multirow{2}{*}{Mental} & \rcells{0.2}{0.3809}{0.6057} \\
      & & \rcells{0.15}{0.3534}{0.4728} \\
    \bottomrule
  \end{tabular}
\end{table}

\begin{table}[ht]
  \centering
  \caption{Unselected Smashed-Data DP operating points.}
  \label{tab:unselected_dp}
  \footnotesize
  \setlength{\tabcolsep}{3.5pt}
  \begin{tabular}{llrrr}
    \toprule
    \textbf{Model} & \textbf{Dataset} & $\sigma$ & \makecell{\textbf{Acc. /}\\\textbf{ROUGE-1}} & \makecell{\textbf{Direct Fwd}\\\textbf{ROUGE-1}} \\
    \midrule
    \multirow{6}{*}{Llama-3.2-3B}
      & \multirow{2}{*}{Banking} & \rycells{0.6}{0.3320}{0.0053} \\
      & & \rycells{1}{0.2979}{0.0485} \\
      \cmidrule{2-5}
      & \multirow{2}{*}{CLINC} & \rycells{0.6}{0.3066}{0.0065} \\
      & & \rycells{1}{0.2422}{0.0389} \\
      \cmidrule{2-5}
      & \multirow{2}{*}{Mental} & \ugcells{0.2}{0.3883}{0.6130} \\
      & & \rycells{1}{0.1119}{0.2614} \\
    \midrule
    \multirow{6}{*}{Llama-3.1-8B}
      & \multirow{2}{*}{Banking} & \gcells{0.54}{0.0000}{0.0063} \\
      & & \gcells{0.87}{0.0000}{0.0109} \\
      \cmidrule{2-5}
      & \multirow{2}{*}{CLINC} & \rycells{0.54}{0.1240}{0.0111} \\
      & & \gcells{0.87}{0.0000}{0.0050} \\
      \cmidrule{2-5}
      & \multirow{2}{*}{Mental} & \ugcells{0.2}{0.3940}{0.5999} \\
      & & \gcells{0.87}{0.0749}{0.2251} \\
    \midrule
    \multirow{6}{*}{Ministral-3-8B}
      & \multirow{2}{*}{Banking} & \gcells{0.6}{0.0000}{0.0026} \\
      & & \gcells{1}{0.0000}{0.0000} \\
      \cmidrule{2-5}
      & \multirow{2}{*}{CLINC} & \gcells{0.6}{0.0000}{0.0032} \\
      & & \gcells{1}{0.0000}{0.0000} \\
      \cmidrule{2-5}
      & \multirow{2}{*}{Mental} & \ugcells{0.2}{0.4087}{0.6820} \\
      & & \rcells{1}{0.3072}{0.2670} \\
    \midrule
    \multirow{6}{*}{Ministral-3-14B}
      & \multirow{2}{*}{Banking} & \rycells{0.6}{0.5977}{0.0124} \\
      & & \rycells{1}{0.5264}{0.0000} \\
      \cmidrule{2-5}
      & \multirow{2}{*}{CLINC} & \rycells{0.6}{0.5986}{0.0173} \\
      & & \rycells{1}{0.3955}{0.0000} \\
      \cmidrule{2-5}
      & \multirow{2}{*}{Mental} & \ugcells{0.2}{0.4073}{0.6878} \\
      & & \rcells{0.6}{0.3684}{0.5811} \\
    \bottomrule
  \end{tabular}
\end{table}

\begin{table}[ht]
  \centering
  \caption{Unselected NoPeek operating points.}
  \label{tab:unselected_nopeek}
  \footnotesize
  \setlength{\tabcolsep}{3.5pt}
  \begin{tabular}{llrrr}
    \toprule
    \textbf{Model} & \textbf{Dataset} & $\lambda$ & \makecell{\textbf{Acc. /}\\\textbf{ROUGE-1}} & \makecell{\textbf{Direct Fwd}\\\textbf{ROUGE-1}} \\
    \midrule
    \multirow{7}{*}{Llama-3.2-3B}
      & \multirow{2}{*}{Banking} & \gcells{$3\times10^{3}$}{0.0000}{0.0479} \\
      & & \gcells{$1\times10^{5}$}{0.0000}{0.0096} \\
      \cmidrule{2-5}
      & \multirow{3}{*}{CLINC} & \gcells{$1\times10^{2}$}{0.0000}{0.0137} \\
      & & \gcells{$3\times10^{3}$}{0.0000}{0.0010} \\
      & & \gcells{$1\times10^{5}$}{0.0000}{0.0000} \\
      \cmidrule{2-5}
      & \multirow{2}{*}{Mental} & \gcells{$3\times10^{3}$}{0.0045}{0.1343} \\
      & & \gcells{$1\times10^{5}$}{0.0000}{0.0000} \\
    \midrule
    \multirow{9}{*}{Llama-3.1-8B}
      & \multirow{4}{*}{Banking} & \ugcells{$2.5\times10^{1}$}{0.7188}{0.1268} \\
      & & \gcells{$1\times10^{2}$}{0.0000}{0.0628} \\
      & & \gcells{$1\times10^{3}$}{0.0000}{0.0031} \\
      & & \gcells{$1\times10^{4}$}{0.0000}{0.0022} \\
      \cmidrule{2-5}
      & \multirow{3}{*}{CLINC} & \ugcells{$5\times10^{1}$}{0.8027}{0.0712} \\
      & & \gcells{$1\times10^{3}$}{0.0000}{0.0265} \\
      & & \gcells{$1\times10^{4}$}{0.0000}{0.0000} \\
      \cmidrule{2-5}
      & \multirow{2}{*}{Mental} & \gcells{$1\times10^{3}$}{0.0687}{0.0544} \\
      & & \gcells{$1\times10^{4}$}{0.0003}{0.0056} \\
    \midrule
    \multirow{11}{*}{Ministral-3-8B}
      & \multirow{4}{*}{Banking} & \gcells{$5\times10^{1}$}{0.0312}{0.0014} \\
      & & \gcells{$1\times10^{2}$}{0.0000}{0.0000} \\
      & & \gcells{$1\times10^{3}$}{0.0000}{0.0000} \\
      & & \gcells{$2\times10^{4}$}{0.0000}{0.0000} \\
      \cmidrule{2-5}
      & \multirow{5}{*}{CLINC} & \gcells{$2.5\times10^{1}$}{0.0449}{0.0159} \\
      & & \gcells{$5\times10^{1}$}{0.0000}{0.0000} \\
      & & \gcells{$1\times10^{2}$}{0.0000}{0.0491} \\
      & & \gcells{$1\times10^{3}$}{0.0000}{0.0000} \\
      & & \gcells{$2\times10^{4}$}{0.0000}{0.0000} \\
      \cmidrule{2-5}
      & \multirow{2}{*}{Mental} & \gcells{$1\times10^{3}$}{0.0153}{0.0644} \\
      & & \gcells{$2\times10^{4}$}{0.0000}{0.0000} \\
    \midrule
    \multirow{6}{*}{Ministral-3-14B}
      & \multirow{2}{*}{Banking} & \gcells{$2\times10^{2}$}{0.0000}{0.0000} \\
      & & \gcells{$3\times10^{3}$}{0.0000}{0.0000} \\
      \cmidrule{2-5}
      & \multirow{2}{*}{CLINC} & \gcells{$2\times10^{2}$}{0.0000}{0.0021} \\
      & & \gcells{$3\times10^{3}$}{0.0000}{0.0006} \\
      \cmidrule{2-5}
      & \multirow{2}{*}{Mental} & \gcells{$2\times10^{2}$}{0.0005}{0.0147} \\
      & & \gcells{$3\times10^{3}$}{0.0000}{0.0500} \\
    \bottomrule
  \end{tabular}
\end{table}

\section{Direct Forward Attack Examples}
\label{sec:appendix_examples}

\begin{figure*}[t]
    \centering
    \includegraphics[width=\linewidth]{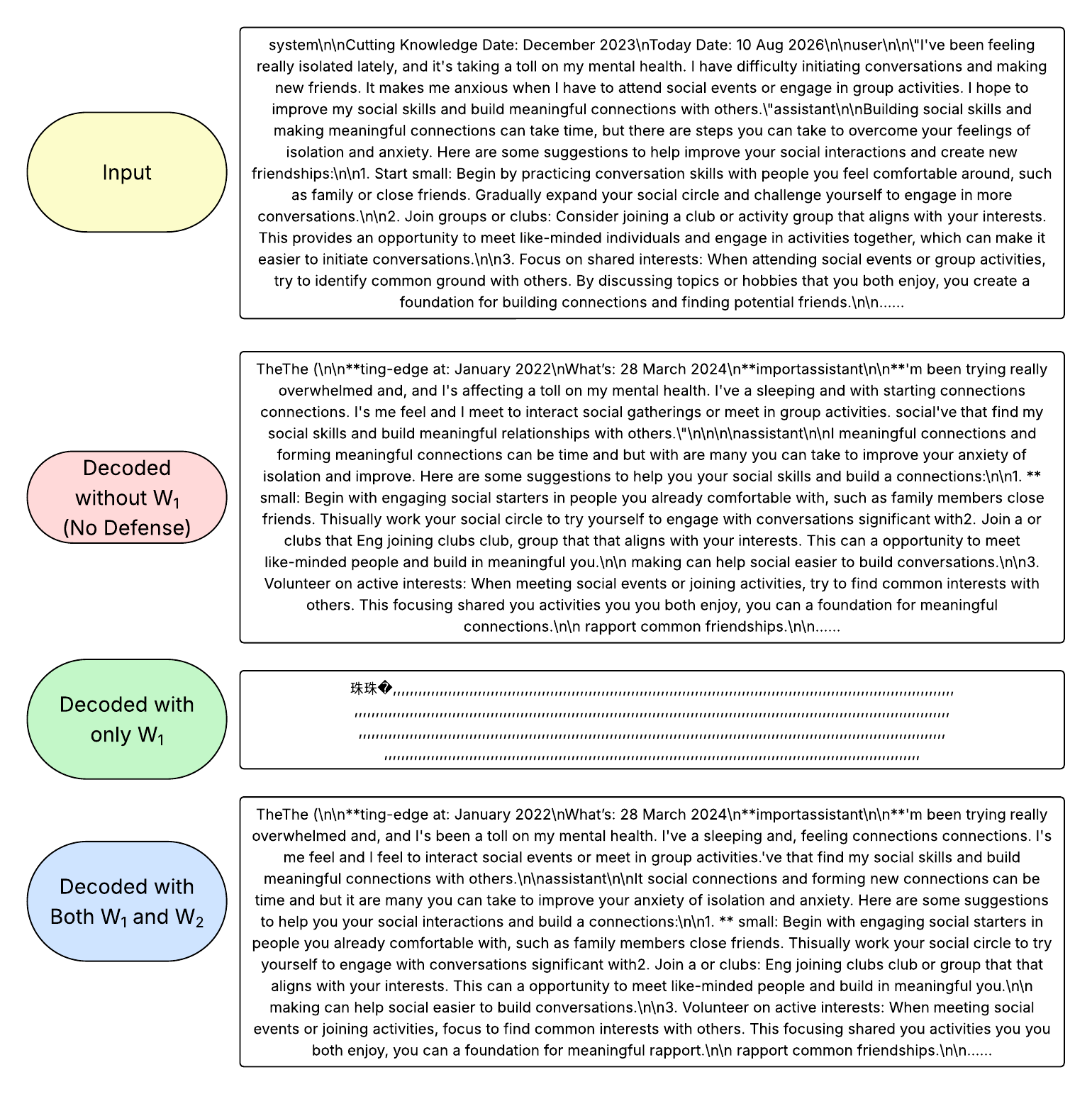}
    \caption{Four text examples from Llama-3.2-3B showing the effect of each defense stage. Each example displays the same input and compares: (1) the true label (ground truth), (2) the server's Direct Forward output without any defense, (3) the server's Direct Forward output when only the $W_1$ obfuscation adapter is applied, and (4) the client's output after applying both $W_1$ and $W_2$.}
    \label{fig:example}
\end{figure*}

Figure~\ref{fig:example} shows qualitative examples on Llama-3.2-3B. Each block shows the true label and compares three outputs for the same input: the server's Direct Forward output without any defense, the server's Direct Forward output when only $W_1$ obfuscation is applied, and the client's output after the full $W_1+W_2$ adapter protocol. Without defense, the server can extract meaningful information from the smashed data. With $W_1$ only, direct decoding by the server becomes uninformative because the obfuscated activations are incompatible with the server's decoder. With both $W_1$ and $W_2$, the client recovers substantially more coherent and task-relevant output from the returned activations while the server observes only the obfuscated path.

% that's all folks
\end{document}